%% file: iclr2025_conference.tex
\documentclass{article} 
\PassOptionsToPackage{table,dvipsnames}{xcolor}
\usepackage{iclr2025_conference,times}

\usepackage{algorithm}
\usepackage{algpseudocode}

\input{math_commands.tex}

\usepackage{hyperref}
\usepackage{url}

\usepackage{placeins}
\usepackage{graphicx}
\usepackage{caption}

\usepackage{bm}
\usepackage{mathrsfs}
\usepackage{booktabs}
\definecolor{Gray}{gray}{0.95}
\definecolor{Green}{rgb}{0.3, 0.73, 0.09}
\definecolor{Red}{rgb}{0.82, 0.1, 0.26}

\definecolor{darkgreen}{rgb}{0.0, 0.5, 0.0}

\title{\textsc{Training} \NoCaseChange{{\normalfont n}}\textsc{GPT}}

\iclrfinalcopy
\author{Ilya Loshchilov \& Boris Ginsburg \\
NVIDIA\\
\texttt{\{iloshchilov,bginsburg\}@nvidia.com} 
}

\begin{document}

\maketitle

\begin{abstract}
The normalized Transformer (nGPT) realizes hyperspherical representation learning by constraining model parameter vectors and activation vectors to the unit hypersphere. In this paper, we describe a practical training recipe for nGPT and evaluate it on modern hybrid Mamba-2--Transformer Mixture-of-Experts (MoE) models. The recipe introduces Logit Gradient Preconditioning, Logarithmic Learning Rate Decay, GatedAdamW, angular update control, and optional exploration mechanisms. Compared with an unnormalized model of the same hybrid MoE architecture trained with AdamW, the 30B-total-parameter nGPT model reaches the same validation loss using approximately half as many training tokens. The recipe scales across the models considered, which contain up to 30B total parameters.
\end{abstract}

\begin{figure}[!b]
    \centering
    \includegraphics[width=0.98\textwidth]{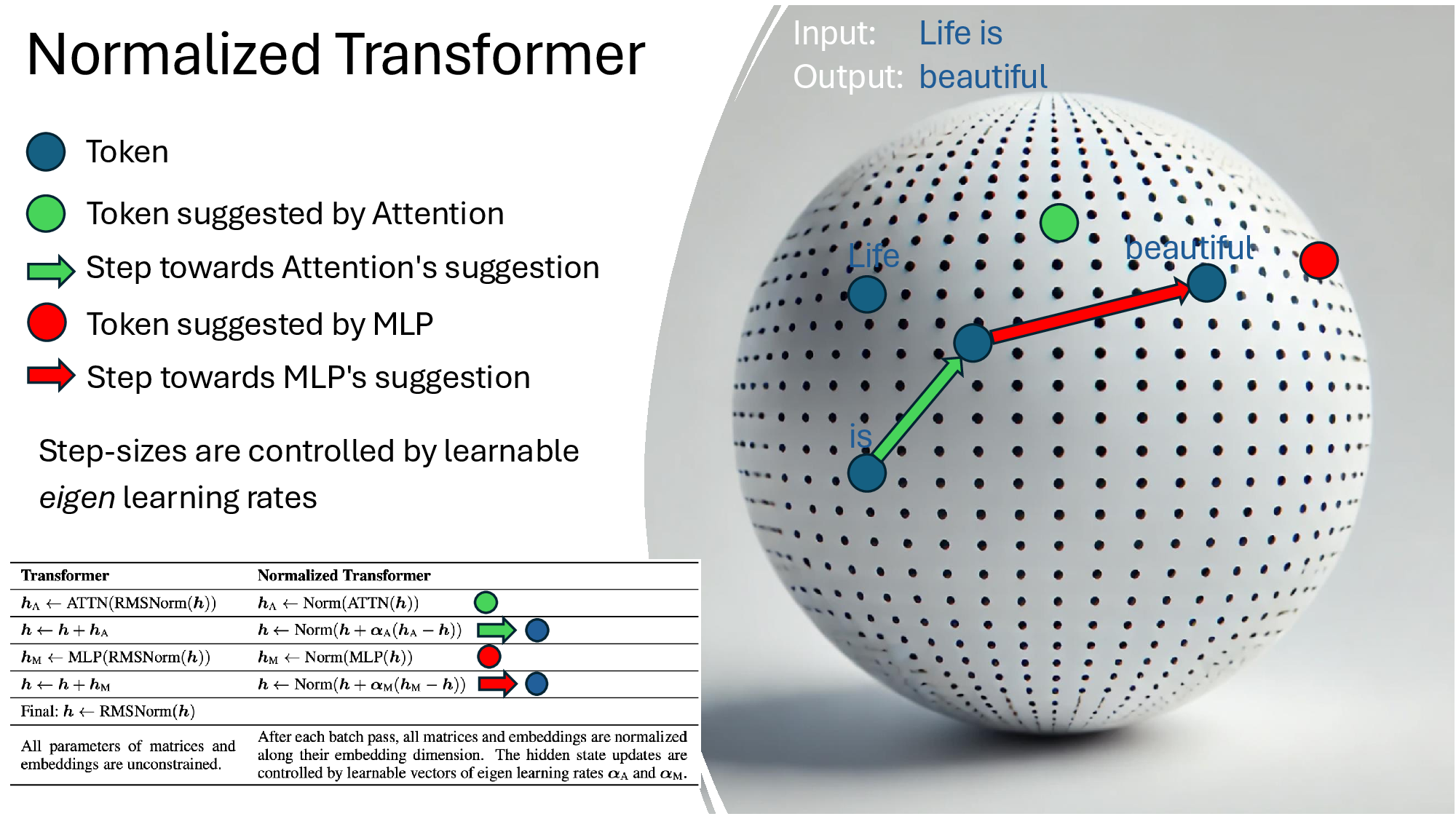}
    \caption{nGPT's forward pass as a multi-step optimization on the hypersphere.}
    \label{fig1:ngpt}
\end{figure}

\section{Preliminaries}

The normalized Transformer (nGPT) paper \citep{loshchilov2024ngpt} proposed a hyperspherical representation in which all activation vectors and parameter vectors that form matrices are normalized to unit norm. The high-level idea behind nGPT is illustrated in Figure~\ref{fig1:ngpt}, where the initial sequence ``Life is'' is used to predict ``beautiful'' via a two-step process within a single layer. The first step of the attention block considers the two blue tokens ``Life'' and ``is'', represented as points, and suggests a prediction depicted by the green point. This prediction is then combined with the hidden state to produce a new intermediate point. The resulting intermediate point is passed to the MLP block, which in turn produces its suggestion depicted by the red point. The hidden
state then takes a second learned step towards the MLP suggestion. When this process is repeated across layers, next-token prediction can be viewed as an optimization process
on the hypersphere: model vectors act as anchors, dot products measure
similarity to these anchors, and the attention and MLP blocks propose
successive steps.

nGPT was designed based on a first-principles view of the hypersphere as the
representation manifold, following an earlier attempt to control the norms of
Transformer parameters during training \citep{loshchilov2023weight}. At the
same time, it connects to a broad line of work that either explicitly studies
hyperspherical representations
\citep{liu2017deep,wang2017normface,liu2018decoupled,xu2018spherical,
mettes2019hyperspherical,wang2020understanding,liu2021learning,karras2024analyzing}
or arrives at related ideas through normalization, decoupled weight decay,
norm control, and rotational training dynamics
\citep{salimans2016weight,loshchilov2017decoupled,franke2023cpr,
kodryan2022training,kosson2023rotational}.

In this work, we extend nGPT from the dense Transformer setting originally
studied in \citet{loshchilov2024ngpt} to modern hybrid Mamba-2--Transformer MoE language
models, using the Nemotron-3 architecture and an industry-grade data blend
\citep{blakeman2025nemotron}. We refer to the complete collection of
normalization, architectural, and optimization modifications considered in
this work as the \emph{nGPT training recipe}. Its principal components include
Logit Gradient Preconditioning, logarithmic learning rate decay, GatedAdamW,
Pre-Moment Tangent Projection, Angular Step Cap, Second-Moment Growth
Clipping, and Post-Moment Exploration Noise. These names refer to individual
components of the recipe rather than to separate model architectures.

The core recipe is not specific to MoE models. For a dense feed-forward block,
the same normalization and optimization rules can be applied to the ordinary
MLP projection matrices, while the router- and expert-specific modifications
are omitted. In the experiments, we use GPT and nGPT to denote, respectively,
the unnormalized and normalized variants of the same hybrid Mamba-2--Transformer MoE architecture; optimizer names are appended only when
the distinction is relevant. The remainder of the paper first introduces the
individual training components and then describes their application to nGPT
in the hybrid MoE setting.

\paragraph{Scope.}
The experiments in this paper were run with a limited compute budget. As a
result, we focus on the training recipe that worked in practice and do not
attempt an exhaustive ablation study. In particular, comprehensive component-wise ablations and hyperparameter scaling laws are left for future work.

\section{Optimization Ingredients}

\subsection{Logit Gradient Preconditioning (LGP)}

In nGPT, the output embedding vectors and hidden states are normalized, so the
unscaled logits are based on bounded dot products. As in the original nGPT
formulation, we therefore use a learnable vocabulary-wise scale vector
$\vs_z \in \mathbb{R}^{V}$ to control the sharpness of the output distribution.
We initialize this vector as
\begin{align}
    \vs_{z,0} = \mathbf{1}.
\end{align}
Let $\vu \in \mathbb{R}^{V}$ denote the logits before this scale is applied.
The forward pass uses
\begin{align}
    \vz = \vs_z \odot \vu,
    \label{eq:sz}
\end{align}
and the cross-entropy loss is computed from $\vz$.

\begin{figure}[!t]
    \centering
    \includegraphics[width=0.8\textwidth]{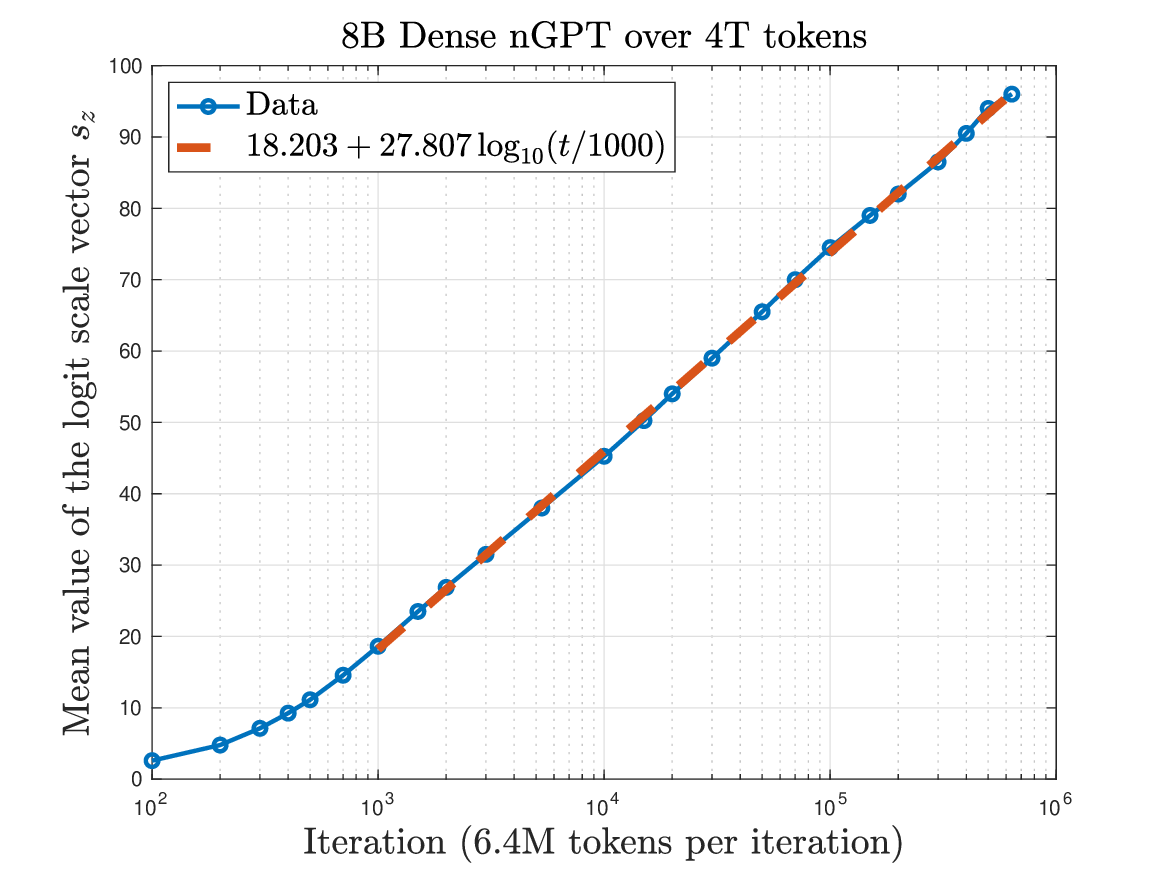}
    \caption{Mean value of the logit scale vector $\vs_z$ for an 8B dense nGPT model trained on 4T tokens. The fit is computed from iteration $t \geq 1000$.}
    \label{fig2:sz}
\end{figure}

The motivation for LGP came from an empirical observation in a dense 8B nGPT
model trained on 4T tokens alongside the
Nemotron-H pretraining cycle \citep{blakeman2025nemotron}.
Figure~\ref{fig2:sz} shows that after an initial transient, the mean value of the learned vocabulary logit scale vector $\vs_z$ grows logarithmically. 
While different token-specific entries of $\vs_z$ grow at different rates, 
the mean of $\vs_z$ appears as an explicit global multiplier on the gradient
propagated through the output layer to all preceding layers, independently of the explicit learning rate schedule. This hidden, time-varying multiplier can complicate the design of optimizers and architectures, and may be one contributor to the
commonly observed growth of gradient norms during long pretraining runs.

Standard backpropagation would multiply the gradient entering the output layer
by $\vs_z$:
\begin{align}
    \frac{\partial \mathcal{L}}{\partial \vu}
    =
    \vs_z \odot
    \frac{\partial \mathcal{L}}{\partial \vz}.
\end{align}
This couples the learned logit scales to the effective learning rate of the
output layer and all preceding layers. In particular, if the mean of
$\vs_z$ grows during training, then the average gradient scale entering the
network also grows, even though this change is not part of the explicit
learning rate schedule. To control this coupling, we use $\vs_z$ in the forward pass but replace its explicit backward multiplier by a normalized scale:
\begin{align}
    \frac{\partial \mathcal{L}}{\partial \vu}
    \leftarrow
    \left(
    \frac{\vs_z}{\mathrm{mean}(\vs_z)}
    \right)^q
    \odot
    \frac{\partial \mathcal{L}}{\partial \vz},
\end{align}
where $q$ controls the strength of the preconditioning. We call this procedure Logit Gradient Preconditioning (LGP), since it
preconditions the gradient flowing backward through the logit scale while
leaving the forward logits unchanged.

In our experiments, we use $q=1$. The two endpoint settings have simple
interpretations. For $q=0$, the explicit backward scale is
\begin{align}
    \left(
        \frac{\vs_z}{\operatorname{mean}(\vs_z)}
    \right)^0
    =
    \mathbf{1},
\end{align}
so the direct multiplicative effect of the learned logit scale on the gradient
propagated from the output layer into the rest of the network is removed.
The scale vector still affects the backward pass indirectly because
$\partial\mathcal{L}/\partial\vz$ depends on the scaled forward logits $\vz$.
For $q=1$, the explicit backward scale is
\begin{align}
    \frac{\vs_z}{\operatorname{mean}(\vs_z)},
\end{align}
which removes the global mean scale while preserving the relative
vocabulary-wise variation in $\vs_z$. Thus, $q=0$ removes the direct
vocabulary-wise multiplier from the output layer Jacobian, whereas $q=1$
retains its relative vocabulary-wise preconditioning effect while removing
the explicit time-varying global multiplier.

The gradient with respect to $\vs_z$ itself is left unchanged and remains the
standard gradient of the forward computation. Therefore, $\vs_z$ remains a learnable vocabulary-wise logit-scale parameter, while its explicit backward effect on the rest of the network is controlled independently. Except when the
replacement scale equals the forward scale, LGP should be viewed as a
preconditioned backward pass rather than the exact gradient of the forward loss
with respect to $\vu$.

For values of $q$ other than zero or one, the mean of
$(\vs_z/\operatorname{mean}(\vs_z))^q$ is not generally equal to one. An
alternative is to normalize after taking the element-wise power by defining
the backward scale for coordinate $i$ as
\begin{align}
    \widetilde b_{q,i}(\vs_z)
    =
    \frac{s_{z,i}^q}
         {\frac{1}{V}\sum_{\ell=1}^{V}s_{z,\ell}^q},
    \qquad i=1,\ldots,V.
\end{align}
This alternative preserves a unit mean backward scale for every $q$ and agrees
with the formulation above at both $q=0$ and $q=1$. We leave the comparison of
the two formulations and the study of intermediate values of $q$ to future
work.  
The element-wise power assumes positive entries of $\vs_z$. We initialize all
entries to one, and the reported experiments use $q=1$, for which no fractional
power is required. Experiments with noninteger values of $q$ should enforce
positivity, for example through a positive parameterization of $\vs_z$.

LGP is not specific to nGPT. In a
standard Transformer, the rows of the output embedding matrix are not
normalized. Decomposing each row as
$\vw_i=s_{z,i}\bar{\vw}_i$, where $s_{z,i}=\|\vw_i\|_2$ and
$\|\bar{\vw}_i\|_2=1$, shows that the row norms act as vocabulary-wise logit
scales. They therefore play the same role as the explicit $\vs_z$ parameter in
nGPT. LGP can in principle be applied to a standard Transformer by computing these
row norms, 
using the equivalent decomposition
$\mW_{\mathrm{out}}=\mathrm{diag}(\vs_z)\bar{\mW}_{\mathrm{out}}$, and replacing
the backward scale associated with $\vs_z$ by
$(\vs_z/\mathrm{mean}(\vs_z))^q$, while leaving the forward logits unchanged.

Preliminary experiments suggest that LGP with $q=1$ removes this explicit,
time-varying global factor without observable performance loss while keeping the forward pass unchanged.

\subsection{Logarithmic Learning Rate Decay}

\begin{figure}[!t]
    \centering
    \includegraphics[width=0.9\textwidth]{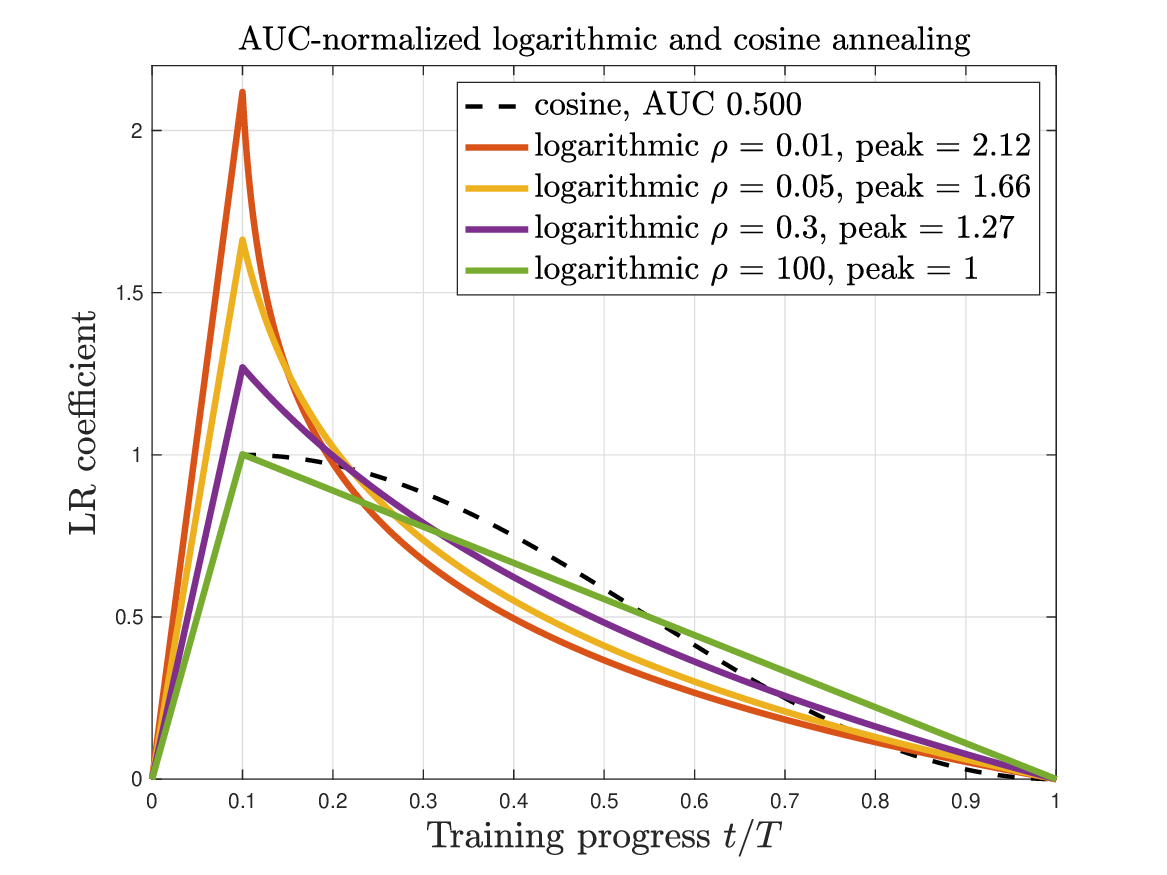}
    \caption{Logarithmic annealing schedules compared with cosine annealing after
matching the area under the curve (AUC). All schedules use a 10\% linear
warmup. Each logarithmic schedule is multiplied by a constant scale factor so
that its AUC matches that of the cosine schedule. Smaller $\rho$ allocates more of the learning rate budget
early, while large $\rho$ approaches linear decay.}
    \label{fig3:logannealing}
\end{figure}

An advantage of nGPT over the standard Transformer parameterization is that
parameter-vector norms are constrained and weight decay is not used for these
vectors. This removes the need to tune the interaction between the learning
rate and weight decay and makes the effective step size more directly
controlled by the learning rate schedule. We therefore consider a more
front-loaded decay schedule, which we call Logarithmic Learning Rate Decay.

After an optional warmup, let
\begin{align}
    r
    =
    \frac{t-t_{\mathrm{warmup}}}
         {t_{\mathrm{decay}}-t_{\mathrm{warmup}}},
    \qquad r\in[0,1],
\end{align}
denote the normalized decay progress. We define a dimensionless schedule multiplier that interpolates between
$\eta_{\max}$ and $\eta_{\min}$ logarithmically in the normalized training
progress: 
\begin{align}
    \eta(t)
    =
    \eta_{\min}
    +
    \left(\eta_{\max}-\eta_{\min}\right)
    \left[
        1
        -
        \frac{
            \log\left(1+\frac{r}{\rho}\right)
        }{
            \log\left(1+\frac{1}{\rho}\right)
        }
    \right],
\end{align}
where $\rho>0$ controls the shape of the decay. At optimizer step $t$, we write $\eta_t=\eta(t)$. The corresponding scheduled
adaptive step size is
\begin{align}
    \alpha_t
    =
    \alpha\,\eta_t,
\end{align}
where $\alpha$ is the base step size in the AdamW notation used below. When
$\eta_{\max}=1$, $\alpha$ is the peak learning rate. The schedule satisfies
$\eta(t_{\mathrm{warmup}})=\eta_{\max}$ and
$\eta(t_{\mathrm{decay}})=\eta_{\min}$. As $\rho\rightarrow\infty$, it
approaches linear decay. Smaller values of $\rho$ produce a faster initial
decrease and a longer tail.

Figure~\ref{fig3:logannealing} compares the proposed logarithmic decay with
cosine decay \citep{loshchilov2016sgdr}. All schedules in this example use a
linear warmup over the first 10\% of training. For this comparison, each
logarithmic schedule is multiplied by a constant so that its area under the
curve (AUC) matches that of the cosine schedule. This keeps the integrated
learning rate budget fixed while changing how that budget is distributed over
training. Smaller values of $\rho$ allocate more of the learning rate budget
early, producing a sharp decrease after warmup and a long tail. Because of the
AUC normalization, the peak schedule multiplier can exceed one.

\subsection{GatedAdamW}

AdamW decouples weight decay from the adaptive gradient update
\citep{loshchilov2017decoupled}. We follow the notation of AdamW and denote by
$\bm{\theta}_t$ the parameters, by $\vg_t$ the stochastic gradient, by
$\vm_t$ and $\vv_t$ the first and second moment estimates, by $\alpha$ the base
step size, by $\eta_t$ the dimensionless schedule multiplier, and by $\lambda$
the decoupled weight decay coefficient. The scheduled adaptive step size is
therefore $\alpha_t=\alpha\eta_t$.

In AdamW, the adaptive update is controlled by
$\hat{\vm}_t/(\sqrt{\hat{\vv}_t}+\epsilon)$. The constant $\epsilon$ is usually
introduced for numerical stability, but it also suppresses updates when
$\sqrt{\hat{\vv}_t}$ is comparable to or smaller than $\epsilon$. We found it
useful to separate these two roles. GatedAdamW uses
$\epsilon_{\mathrm{num}}$ for numerical stability and a separate soft gate to
control the update applied at small second-moment scales.

Let
\begin{align}
    \vd_t
    =
    \sqrt{\hat{\vv}_t} + \epsilon_{\mathrm{num}},
\end{align}
where $\epsilon_{\mathrm{num}}\geq 0$ is a numerical constant. We define the
coordinate-wise gate
\begin{align}
    \bm{\gamma}_t
    =
    \sigma\left(
        a \log \frac{\vd_t}{\epsilon_{\mathrm{gate}}}
    \right),
\end{align}
where $\sigma(\cdot)$ is the sigmoid function,
$\epsilon_{\mathrm{gate}}>0$ is the gate threshold, and $a>0$ controls the
sharpness of the gate. The GatedAdamW update is then
\begin{align}
    \bm{\theta}_t
    =
    \bm{\theta}_{t-1}
    -
    \eta_t
    \left(
        \alpha\,
        \bm{\gamma}_t
        \odot
        \frac{\hat{\vm}_t}{\vd_t}
        +
        \lambda \bm{\theta}_{t-1}
    \right).
\end{align}

The gate has a simple interpretation. Coordinates with
$\vd_{t,i} \ll \epsilon_{\mathrm{gate}}$ receive a small gate value, which
suppresses the numerically normalized direction $\hat m_{t,i}/d_{t,i}$.
Because $\epsilon_{\mathrm{num}}$ may be smaller than the AdamW epsilon, the
resulting update can nevertheless be larger than the corresponding AdamW
update, particularly when $a<1$. Coordinates with
$\vd_{t,i} \gg \epsilon_{\mathrm{gate}}$ receive a gate value close to one and
experience little gate suppression. Thus, GatedAdamW makes the effect of
Adam's $\epsilon$ explicit and tunable:
$\epsilon_{\mathrm{num}}$ controls numerical stability, while
$\epsilon_{\mathrm{gate}}$ determines the second-moment scale at which the
gate transitions from suppressed to active updates.

\DeclareRobustCommand{\gatehltext}[1]{%
  \begingroup
  \setlength{\fboxsep}{1.3pt}%
  \colorbox{Green!65}{\strut #1}%
  \endgroup
}

\makeatletter
\newcommand{\gatehlmath}[1]{\mathpalette\gatehlmath@{#1}}
\newcommand{\gatehlmath@}[2]{%
  \begingroup
  \setlength{\fboxsep}{1.3pt}%
  \colorbox{Green!65}{$\m@th#1#2$}%
  \endgroup
}
\makeatother

\begin{algorithm}[t]
\caption[GatedAdamW]{\gatehltext{Gated}AdamW}
\label{alg:gated_adamw}
\begin{algorithmic}[1]
\Require base step size $\alpha$, dimensionless schedule multiplier $\eta_t$, 
         moment coefficients $\beta_1,\beta_2 \in [0,1)$,
         numerical constant $\epsilon_{\mathrm{num}}\geq 0$,
         gate threshold $\epsilon_{\mathrm{gate}}>0$,
         gate sharpness $a>0$,
         weight decay $\lambda$
\Require initial parameters $\bm{\theta}_0$
\State $\vm_0 \gets 0$, $\vv_0 \gets 0$
\For{$t = 1,2,\ldots,T$}
    \State $\vg_t \gets \nabla f_t(\bm{\theta}_{t-1})$
    \State $\vm_t \gets
        \beta_1 \vm_{t-1} + (1-\beta_1)\vg_t$
    \State $\vv_t \gets
        \beta_2 \vv_{t-1} + (1-\beta_2)\vg_t^2$
    \State $\hat{\vm}_t \gets \vm_t / (1-\beta_1^t)$
    \State $\hat{\vv}_t \gets \vv_t / (1-\beta_2^t)$

    \State $\gatehlmath{
        \vd_t \gets
        \sqrt{\hat{\vv}_t} + \epsilon_{\mathrm{num}}
    }$

    \State $\gatehlmath{
        \bm{\gamma}_t \gets
        \sigma\!\left(
            a \log\left(
                \vd_t / \epsilon_{\mathrm{gate}}
            \right)
        \right)
    }$

    \State $\bm{\theta}_t \gets
        \bm{\theta}_{t-1}
        -
        \eta_t
        \left(
            \alpha\,
            \gatehlmath{\bm{\gamma}_t}
            \odot
            \frac{\hat{\vm}_t}{\vd_t}
            +
            \lambda \bm{\theta}_{t-1}
        \right)$
\EndFor
\end{algorithmic}
\end{algorithm}

\begin{figure}[!t]
    \centering
    \includegraphics[width=0.9\textwidth]{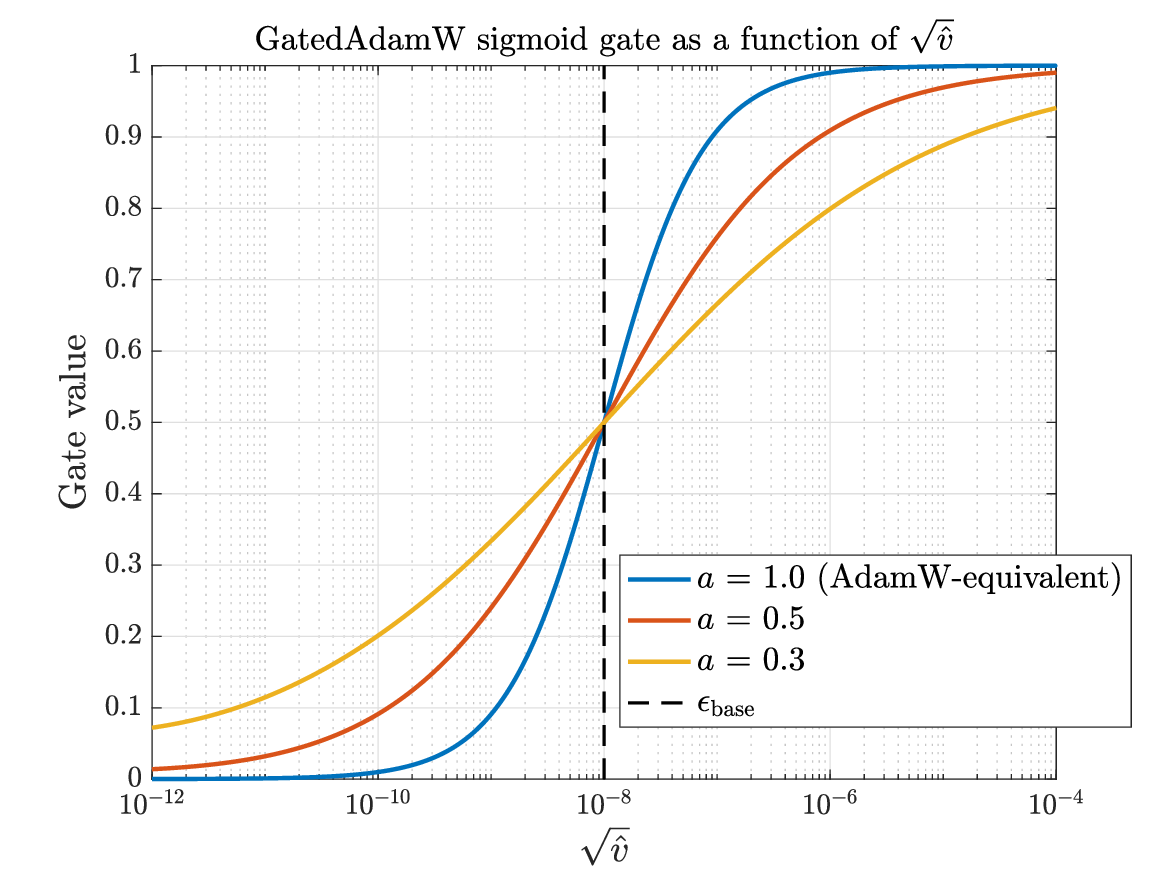}
    \caption{Effect of the GatedAdamW sharpness parameter $a$ on the
coordinate-wise sigmoid gate. The gate is plotted as a function of the
bias-corrected second-moment scale $\sqrt{\hat v}$. The vertical dashed line
marks the gate threshold $\epsilon_{\mathrm{gate}}=10^{-8}$, chosen equal to
the AdamW value of $\epsilon$. All curves cross $0.5$ when
$\sqrt{\hat v}+\epsilon_{\mathrm{num}}=\epsilon_{\mathrm{gate}}$. Smaller
values of $a$ produce a smoother transition, suppressing coordinates below the
threshold less strongly while approaching one more slowly above the
threshold.}
    \label{fig:gatedadamw}
\end{figure}

A useful special case recovers AdamW exactly. If
$a=1$, $\epsilon_{\mathrm{num}}=0$, and
$\epsilon_{\mathrm{gate}}=\epsilon$, then
\begin{align}
    \sigma\left(
        \log \frac{\sqrt{\hat{\vv}_t}}{\epsilon}
    \right)
    =
    \frac{\sqrt{\hat{\vv}_t}}
         {\sqrt{\hat{\vv}_t}+\epsilon}.
\end{align}

For positive $\sqrt{\hat{\vv}_t}$, this gives
\begin{align}
    \bm{\gamma}_t
    \odot
    \frac{\hat{\vm}_t}{\sqrt{\hat{\vv}_t}}
    =
    \frac{\hat{\vm}_t}
         {\sqrt{\hat{\vv}_t}+\epsilon}.
\end{align}
The equality at zero second moment is understood through the corresponding
continuous extension. The update therefore reduces to AdamW. This makes
GatedAdamW a conservative extension of AdamW: this special setting recovers
AdamW exactly, while other settings control the transition between suppressed
and fully active coordinates. An equivalent formulation of the gate without
an explicit sigmoid is given in Appendix~\ref{appendix:equivalentform}.

Figure~\ref{fig:gatedadamw} visualizes the role of the sharpness parameter
$a$ in GatedAdamW. In the figure, we use
$\epsilon_{\mathrm{gate}}=10^{-8}$ and
$\epsilon_{\mathrm{num}}=10^{-14}$. The gate is equal to $0.5$ when
$d_{t,i}=\epsilon_{\mathrm{gate}}$, or equivalently when
$\sqrt{\hat v_{t,i}}=10^{-8}-10^{-14}$, which is visually indistinguishable
from $10^{-8}$ on the plot.

The parameter $a$ controls the sharpness of the transition. When $a=1$ and
$\epsilon_{\mathrm{num}}=0$, the gate exactly recovers the implicit AdamW
epsilon gate. With the small nonzero value of
$\epsilon_{\mathrm{num}}$ used in the figure, the difference is negligible on
the displayed scale. Smaller values of $a$ make the gate softer: they assign
larger gate values to coordinates below the threshold but approach one more
slowly above it. Thus, $a$ controls how abruptly coordinates transition from
the epsilon-dominated regime to receiving the full adaptive update.

\paragraph{Angular Step Cap (ASC).}
We optionally cap the angular displacement of normalized non-embedding
parameter vectors. This provides a safeguard on the geometric step represented
by the candidate after normalization. Let $\vw_{t-1}$ be the previous
normalized vector and let $\widetilde{\vw}_t$ be the post-update candidate
before final normalization, including any optional post-update perturbation.
We extract its component tangent to $\vw_{t-1}$:
\begin{align}
    \vu_t
    =
    \widetilde{\vw}_t
    -
    \vw_{t-1}
    \frac{\langle \vw_{t-1},\widetilde{\vw}_t\rangle}
         {\langle \vw_{t-1},\vw_{t-1}\rangle}.
\end{align}
The angle between the previous vector and the candidate is
\begin{align}
    \phi_t
    =
    \operatorname{atan2}
    \left(
        \|\vu_t\|_2,
        \left\langle
            \frac{\vw_{t-1}}{\|\vw_{t-1}\|_2},
            \widetilde{\vw}_t
        \right\rangle
    \right).
\end{align}
If $\phi_t$ exceeds a cap $\theta_{\max}(t)$ and
$\|\vu_t\|_2>0$, we replace the candidate by the point at angle
$\theta_{\max}(t)$ in the same tangent direction:
\begin{align}
    \widetilde{\vw}_t
    \leftarrow
    \cos\!\left(\theta_{\max}(t)\right)
    \frac{\vw_{t-1}}{\|\vw_{t-1}\|_2}
    +
    \sin\!\left(\theta_{\max}(t)\right)
    \frac{\vu_t}{\|\vu_t\|_2}.
\end{align}
Otherwise, the candidate is left unchanged. In both cases, the usual nGPT
normalization is subsequently applied as the retraction step.

We also propose to optionally use the following techniques in GatedAdamW: 
\textbf{Pre-Moment Tangent Projection} (Appendix \ref{appendix:projection}), 
\textbf{Post-Moment Exploration Noise (PMEN)} (Appendix \ref{appendix:noise}), 
\textbf{Second-Moment Growth Clipping (SMGC)} (Appendix \ref{appendix:clipping}). 
They are described in the Appendix because they are not central to the paper and their impact on the validation loss is modest. 

\newcommand{\vxi}{\boldsymbol{\xi}}

\section{Training {\normalfont n}GPT}

In this section, we recall the principal design choices of nGPT and describe
the modifications used to train it with modern MoE models such as Nemotron-3
\citep{blakeman2025nemotron,waleffe2024empirical}.

\subsection{Normalization and scaling factors}

In nGPT, we constrain all parameter vectors that form matrices to have unit
norm by normalizing them along the embedding dimension
(e.g., $d_{\text{model}}=1024$). In distributed settings, normalization must
be applied to the parameters on which the optimizer operates. A common
implementation bug is to normalize the instantiated model parameters while
leaving the optimizer parameters unconstrained. 

Activations $\vh$ are also normalized to unit norm when they are recombined
with the stream:
\begin{align}
    \vh
    &\leftarrow
    \operatorname{Norm}\!\left(
        \vh
        +
        \bm{\alpha_{\text{A}}}
        \odot
        (\vh_{\text{A}}-\vh)
    \right),
    \label{eq:ATTNnew2}
    \\
    \vh
    &\leftarrow
    \operatorname{Norm}\!\left(
        \vh
        +
        \bm{\alpha_{\text{M}}}
        \odot
        (\vh_{\text{M}}-\vh)
    \right),
    \label{eq:MLPnew2}
\end{align}
where
$\bm{\alpha_{\text{A}}}\in\mathbb{R}_{\geq 0}^{d_{\text{model}}}$ and
$\bm{\alpha_{\text{M}}}\in\mathbb{R}_{\geq 0}^{d_{\text{model}}}$ are
learnable parameters, called eigen learning rates, applied to the
unit-normalized outputs of the attention/state-space model (SSM) and MLP/MoE blocks,
$\vh_{\text{A}}=\operatorname{Norm}(\operatorname{ATTN}(\vh))$ and
$\vh_{\text{M}}=\operatorname{Norm}(\operatorname{MLP}(\vh))$, respectively.

The presence of nonlinear elements in the network may render products of
normalized vectors too constrained. nGPT introduced a trainable vector
$\vs_{qk}\in\mathbb{R}^{d_k}$ to rescale normalized queries $\vq$
and keys $\vk$ (see QKNorm
\citep{henry2020querykey,nguyen2019transformers}):
\begin{align}
    \vq
    &\leftarrow
    \operatorname{Norm}(\vq)\odot\vs_{qk},
    \label{eq:qkscaling1}
    \\
    \vk
    &\leftarrow
    \operatorname{Norm}(\vk)\odot\vs_{qk}.
    \label{eq:qkscaling2}
\end{align}

Similarly, the intermediate activations $\vu$ of the MLP/MoE block are
rescaled by a trainable vector
$\vs_u\in\mathbb{R}^{d_{\text{MLP}}}$:
\begin{align}
    \vu
    \leftarrow
    \vu\odot\vs_u.
    \label{eq:mlpscaling1}
\end{align}

This may be particularly important for activations such as GELU or SwiGLU,
whose inputs must be at an appropriate scale to preserve nonlinearity. 
For squared ReLU, the role of this scale differs from that for saturating or
gated nonlinearities: positive rescaling does not change the activation 
support, but it changes the output magnitude quadratically.

For any trainable vector of scaling parameters such as $\vs_a$, nGPT uses two
scalars, $s_{a,\mathrm{init}}$ and $s_{a,\mathrm{scale}}$. Each entry of
$\vs_a$ is initialized to $s_{a,\mathrm{scale}}$, while the forward pass uses
the effective scaling vector
\begin{align}
    \vs_a^{\mathrm{eff}}
    =
    \vs_a
    \frac{s_{a,\mathrm{init}}}{s_{a,\mathrm{scale}}}.
\end{align}
This allows us to control the effective learning rate of $\vs_a^{\mathrm{eff}}$ by adjusting
$s_{a,\mathrm{scale}}$ while keeping the global learning rate unchanged.

\subsection{Changes specific to Mamba-2 and MoE}

We investigated various normalization options for Mamba-2 components
\citep{dao2024mamba2} and found that both its input and output projections can
be normalized. After removing RMSNorm from the Mamba-2 block, we introduce a
trainable scalar $s_{\mathrm{mamba}}$ to rescale the input activations and
place the inputs to SiLU at an appropriate scale. Similarly, we introduce a
trainable scalar $s_{\mathrm{moe}}$ for the MoE block to place the inputs to
the sigmoid router at an appropriate scale. In both cases, vectors, as in
RMSNorm, could be used instead of scalars. However, we did not observe a
significant difference between the two choices. The scalar choice preserves an
isotropic scaling interpretation within the hyperspherical view. The vector
choice is also compatible with this view and can be interpreted as a diagonal
preconditioner.

\subsection{Summary of modifications}
\label{section_summary}

The recipe for converting the baseline hybrid Mamba-2--Transformer MoE model into its normalized version is as follows:

\begin{enumerate}
    \item Remove normalization layers such as RMSNorm and LayerNorm. Remove
weight decay.
    \item Use GatedAdamW without weight decay (thus, GatedAdam). Apply
Pre-Moment Tangent Projection to parameter vectors selected for normalization.
Set the gating hyperparameter $a$ to 0.5; with the Adam-equivalent epsilon
setting, $a=1$ recovers Adam. Apply an Angular Step Cap
$\theta_{\max}(t)$ to normalized non-embedding vectors, with an optional
warmup over the first $t_{\mathrm{warmup}}$ iterations. After each GatedAdam
update, normalize all selected rows or columns of the input and output
embedding, attention, MoE, router, and Mamba projection matrices. Parameters
not selected for normalization are trained with Adam (GatedAdam with $a=1$). 
    \item Change the softmax scaling factor in attention from
$1/\sqrt{d_k}$ to $\sqrt{d_k}$. Normalize and rescale $\vq$ and $\vk$ as in
Equations~\ref{eq:qkscaling1} and~\ref{eq:qkscaling2}, with
$s_{qk,\mathrm{init}}=s_{qk,\mathrm{scale}}=1$.
    \item Implement the rescaling of the intermediate state of the MoE/MLP block using Equation~\ref{eq:mlpscaling1}, where $\vs_{u}$  uses $s_{u,init}=1$ and $s_{u,scale}=1$.
    \item Implement activation normalization for each recombination as in Equations~\ref{eq:ATTNnew2} and \ref{eq:MLPnew2} with  ${\alpha}_{\text{A},init}={\alpha}_{\text{M},init}=0.1$ (which may be on the order of $1/n_{\text{layers}}$) and ${\alpha}_{\text{A},scale}={\alpha}_{\text{M},scale}=1$.
    \item Implement activation scaling for Mamba-2 with $s_{\text{mamba},init}=0.5\sqrt{d_{\text{model}}}$ and $s_{\text{mamba},scale}=1$ and for MoE with $s_{\text{moe},init}=0.5\sqrt{d_{\text{model}}}$ and $s_{\text{moe},scale}=1$.
    \item Implement the rescaling of logits using Equation~\ref{eq:sz} with $s_{z,init}=1$ and $s_{z,scale}=1$.
    \item Apply Logit Gradient Preconditioning with $q=1$. 
    \item Apply Post-Moment Exploration Noise using the hard-threshold profile at each iteration starting from the first one,
$\tau_v=10^{-10}$, $c_{j,t}=1/\sqrt{d_j}$, and
$\lambda_{\mathrm{noise}}=10$.  
    \item Apply Second-Moment Growth Clipping after the first 500 iterations, with $R=100$, $s_{\min}=10^{-10}$.
    \item In contrast to the original nGPT parameterization, we change
$s_{z,\mathrm{scale}}$, $\alpha_{\mathrm{A},\mathrm{scale}}$,
$\alpha_{\mathrm{M},\mathrm{scale}}$, and $s_{qk,\mathrm{scale}}$ from
$1/\sqrt{d_{\mathrm{model}}}$ to $1$. To compensate for this
reparameterization, we set their effective peak learning rates to $0.5$,
$0.2$, $0.2$, and $0.2$, respectively. In our experiments, we did not find
it necessary to scale these learning rates with model size.
\end{enumerate}

\section{Experiments}

\subsection{Experimental Setup}

All experiments use models and data developed by the Nemotron Team
\citep{blakeman2025nemotron}. More specifically, we use the scaling ladder
developed by \citet{name2026personal}, which consists of Nemotron-3 Nano models
with progressively increasing widths and depths. The model labels refer to
total parameter counts. We consider the following models: 1B (0.21B active
parameters per token, 88B training tokens), 2B (0.37B active, 140B training
tokens), 4B (0.61B active, 215B training tokens), 7B (0.91B active, 310B
training tokens), 14B (1.74B active, 560B training tokens), and 30B (3.23B active, 1000B training tokens). Thus, the
number of training tokens is approximately 320--420 times the number of active
parameters, excluding the input embeddings \citep{name2026personal}. The largest model corresponds to Nemotron-3 Nano. 
Additional details of the ladder models are omitted because they have not yet
been fully described by their authors.

\begin{figure}[!t]
    \centering
    \includegraphics[width=0.98\textwidth]{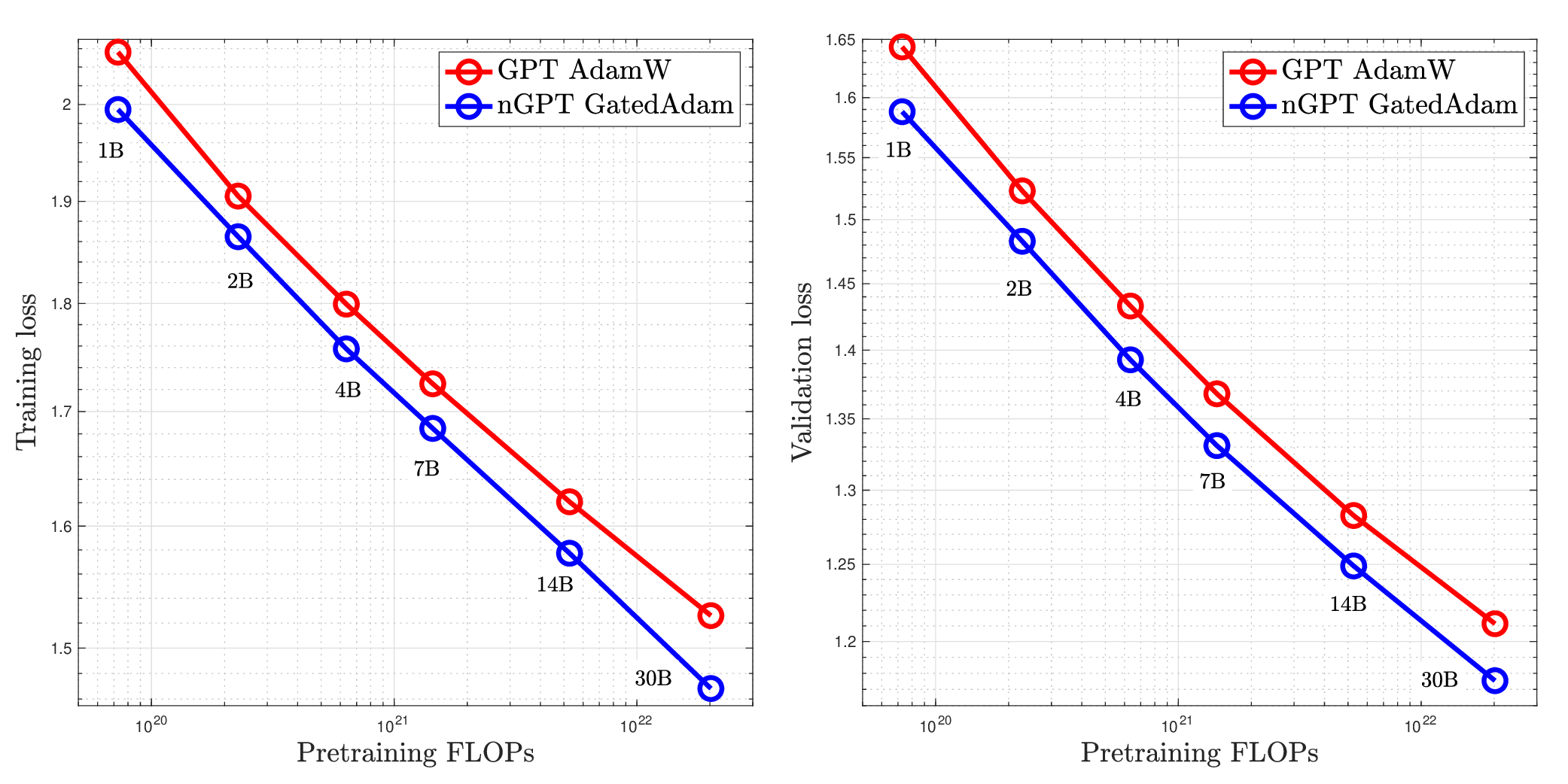}
    \caption{Scaling results for hybrid Mamba-2--Transformer MoE models trained with AdamW (denoted as GPT AdamW) and their normalized versions trained with GatedAdam (denoted as nGPT GatedAdam). Six model sizes (by total number of parameters) are considered: 1B, 2B, 4B, 7B, 14B, and 30B \citep{name2026personal}. The 30B nGPT model trained with GatedAdam incurs about 6\% overhead; one could therefore shift the corresponding nGPT results by 6\% along the x-axis. We do not make this adjustment because the measured overhead is not necessarily attributable to increased FLOPs.}
    \label{fig1:ladder}
\end{figure}

\begin{table}
\centering
\caption{Scaling results for hybrid Mamba-2--Transformer MoE models trained with AdamW (denoted as GPT AdamW) and their normalized versions trained with GatedAdam (denoted as nGPT GatedAdam). Six model sizes (by total number of parameters) are considered: 1B, 2B, 4B, 7B, 14B, and 30B \citep{name2026personal}. }
\label{tab:gpt_ngpt_gatedadam_losses}
\begin{tabular}{rccccccccc}
\toprule
Model size
& Tokens
& \multicolumn{4}{c}{Training loss}
& \multicolumn{4}{c}{Validation loss} \\
\cmidrule(lr){3-6}
\cmidrule(lr){7-10}
&
& GPT
& nGPT
& Diff.
& Diff. \%
& GPT
& nGPT
& Diff.
& Diff. \% \\
\midrule
1B  & 88B
& 2.0561 & 1.9946 & 0.0615 & 2.99\%
& 1.6435 & 1.5881 & 0.0554 & 3.37\% \\

2B  & 140B
& 1.9053 & 1.8648 & 0.0405 & 2.13\%
& 1.5230 & 1.4829 & 0.0401 & 2.63\% \\

4B  & 215B
& 1.7994 & 1.7573 & 0.0421 & 2.34\%
& 1.4330 & 1.3928 & 0.0402 & 2.81\% \\

7B  & 310B
& 1.7251 & 1.6849 & 0.0402 & 2.33\%
& 1.3680 & 1.3310 & 0.0370 & 2.70\% \\

14B & 560B
& 1.6207 & 1.5772 & 0.0435 & 2.69\%
& 1.2827 & 1.2489 & 0.0338 & 2.63\% \\

30B & 1000B
& 1.5260 & 1.4682 & 0.0577 & 3.78\%
& 1.2113 & 1.1754 & 0.0359 & 2.97\% \\
\bottomrule
\end{tabular}
\end{table}

For readability, we refer to a Nemotron-3-style hybrid Mamba--Transformer
model as GPT and to its normalized counterpart as nGPT. The baseline model and
results were provided by \citet{name2026personal}, together with the following
recommended hyperparameters: $\beta_1=0.9$, $\beta_2=0.95$, and weight decay
of $0.1$. After a warmup over 1B tokens, the peak learning rate was set to
$2.2\times10^{-3}$ for the 1B model, $2.0\times10^{-3}$ for the 2B model,
$1.8\times10^{-3}$ for the 4B model, $1.6\times10^{-3}$ for the 7B model, 
$1.4\times10^{-3}$ for the 14B model, and
$1.2\times10^{-3}$ for the 30B model. The learning rate followed the Warmup-Stable-Decay (WSD)
schedule \citep{xing2018walk,hu2024minicpm} and decayed to 1\% of its peak
value. Training sequence length is 8192. The global batch size is fixed for all scales at 768.

For nGPT, we use the proposed logarithmic decay schedule with $\rho=0.05$
(see Figure~\ref{fig3:logannealing}). The learning rates of all parameters
follow this global decay schedule, subject to the parameter-specific
multipliers and warmups described below. In particular, (i) the learning rates
of $\vs_z$ and all Mamba parameters except the output projections are warmed up
during the first 10\% of training, and (ii) the Angular Step Cap for all
normalized non-embedding vectors is increased from $0^\circ$ to $1.0^\circ$
over the same period. We use these warmups in all reported nGPT runs. Their endpoint and duration
were selected in smaller-scale experiments and were not systematically
retuned jointly with the global and parameter-specific learning rates for 
the present scaling ladder.

The peak learning rate for nGPT is set to
$C/\sqrt{d_{\text{model}}}$ (with $C=0.24$ for 1B, 2B, 4B; $C=0.22$ for 7B; $C=0.18$ for 14B; $C=0.12$ for 30B) and subsequently decays to zero. For parameter
vectors in the routed and shared expert matrices, we multiply the learning
rate by $1.5$, while for router parameter vectors, we multiply it by $2.0$.
These multipliers were selected based on experiments with smaller models.
They may be unnecessary; for example, some of their benefit might be recovered
by adjusting the global peak learning rate. However, computational constraints
did not allow us to test this possibility at scale. The baseline GPT uses
$\epsilon=10^{-8}$. For GatedAdam, we set
$\epsilon_{\mathrm{gate}}=10^{-8}$ and
$\epsilon_{\mathrm{num}}=10^{-14}$, and use
$\beta_1=\beta_2=0.975$.

We enable Second-Moment Growth Clipping after the first 500 optimizer steps and
set $R=100$. This deliberately permissive threshold is intended to suppress
only extreme gradient spikes while leaving typical gradients unchanged. In our
experiments, values as low as $R=2$, as well as disabling clipping altogether,
resulted in comparable final performance. We therefore regard this mechanism
primarily as a conservative safeguard against rare optimization instabilities
rather than as an essential component of the training recipe.

\begin{figure}[!t]
    \centering
    \includegraphics[width=0.99\textwidth]{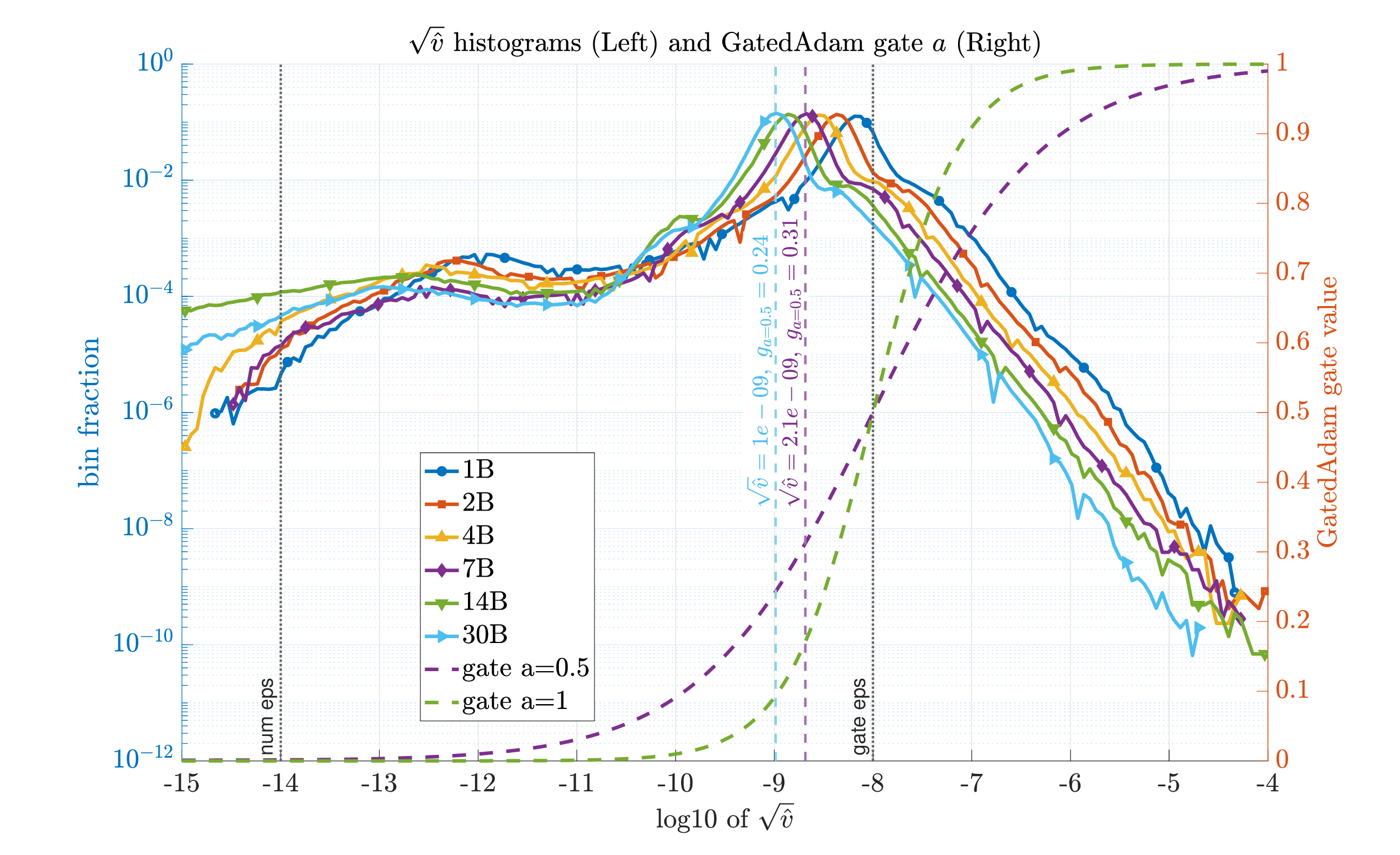}
    \caption{
Histogram of second-moment scales and the corresponding GatedAdam gate.
Solid curves show the distributions of $\log_{10}\sqrt{\hat v}$ across optimizer
parameters for different model sizes using the left logarithmic axis.
Dashed curves show the coordinate-wise GatedAdam gate on the right axis,
$g_a(s)=\sigma\!\left(a\left[\log(s+\epsilon_{\mathrm{num}})
-\log(\epsilon_{\mathrm{gate}})\right]\right)$, where
$s=\sqrt{\hat v}$. Vertical reference lines mark
$\epsilon_{\mathrm{gate}}$, $\epsilon_{\mathrm{num}}$, and the medians of the
distributions. The figure shows which parts of the optimizer state lie in the
$\epsilon$-dominated regime and how the choice of $a$ changes the transition
from suppressed to fully active updates.
}
    \label{fig2:hist}
\end{figure}

\begin{figure}[!t]
    \centering
    \includegraphics[width=0.99\textwidth]{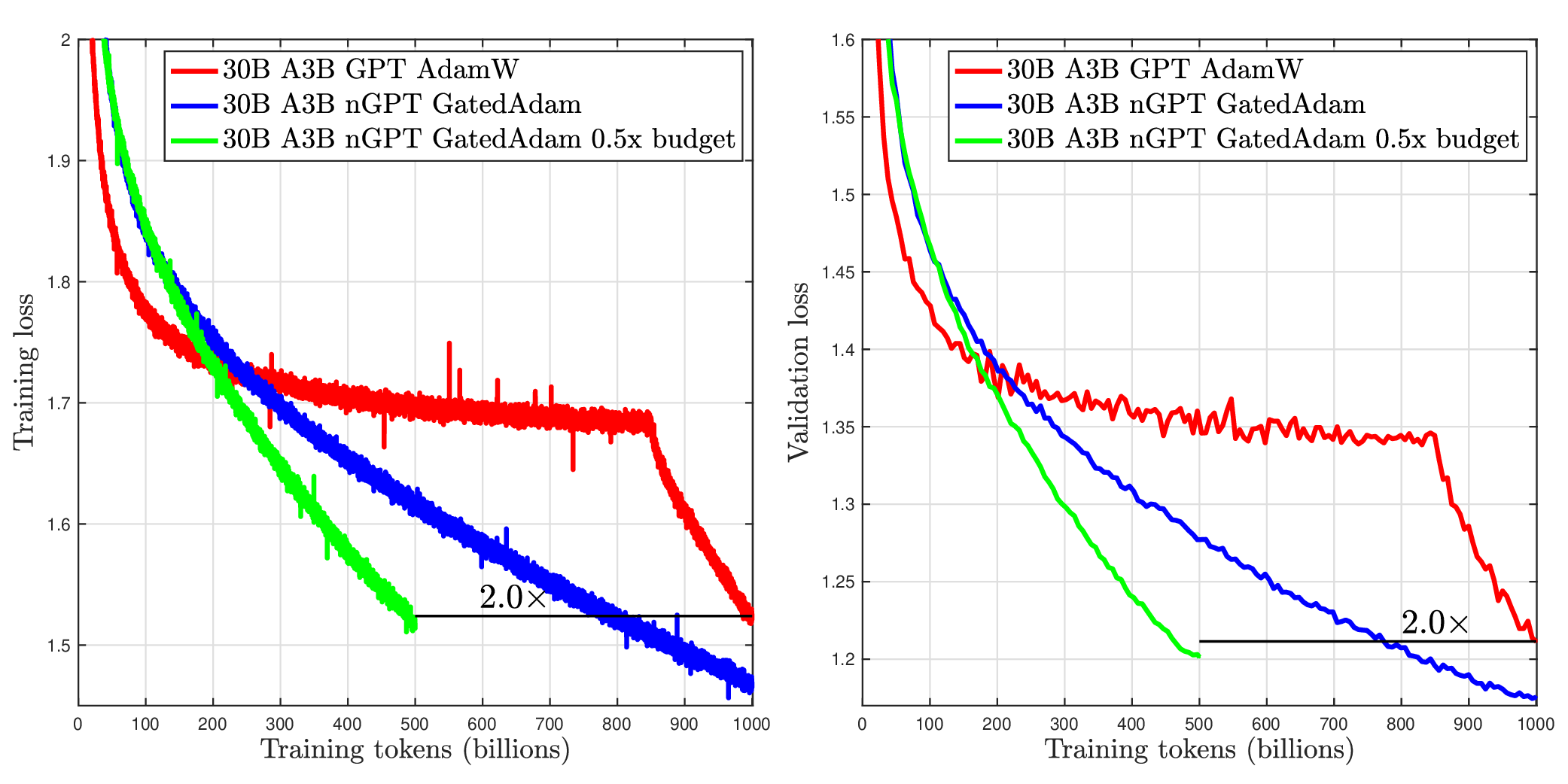}
    \caption{Training (left) and validation (right) losses as functions of the number of training tokens for the MoE model with 30B total parameters and 3.23B active parameters per token. We compare the GPT baseline trained with AdamW and its normalized nGPT counterpart trained with GatedAdam.}
    \label{fig:lossplot}
\end{figure}

\begin{figure}[!t]
    \centering
    \includegraphics[width=0.8\textwidth]{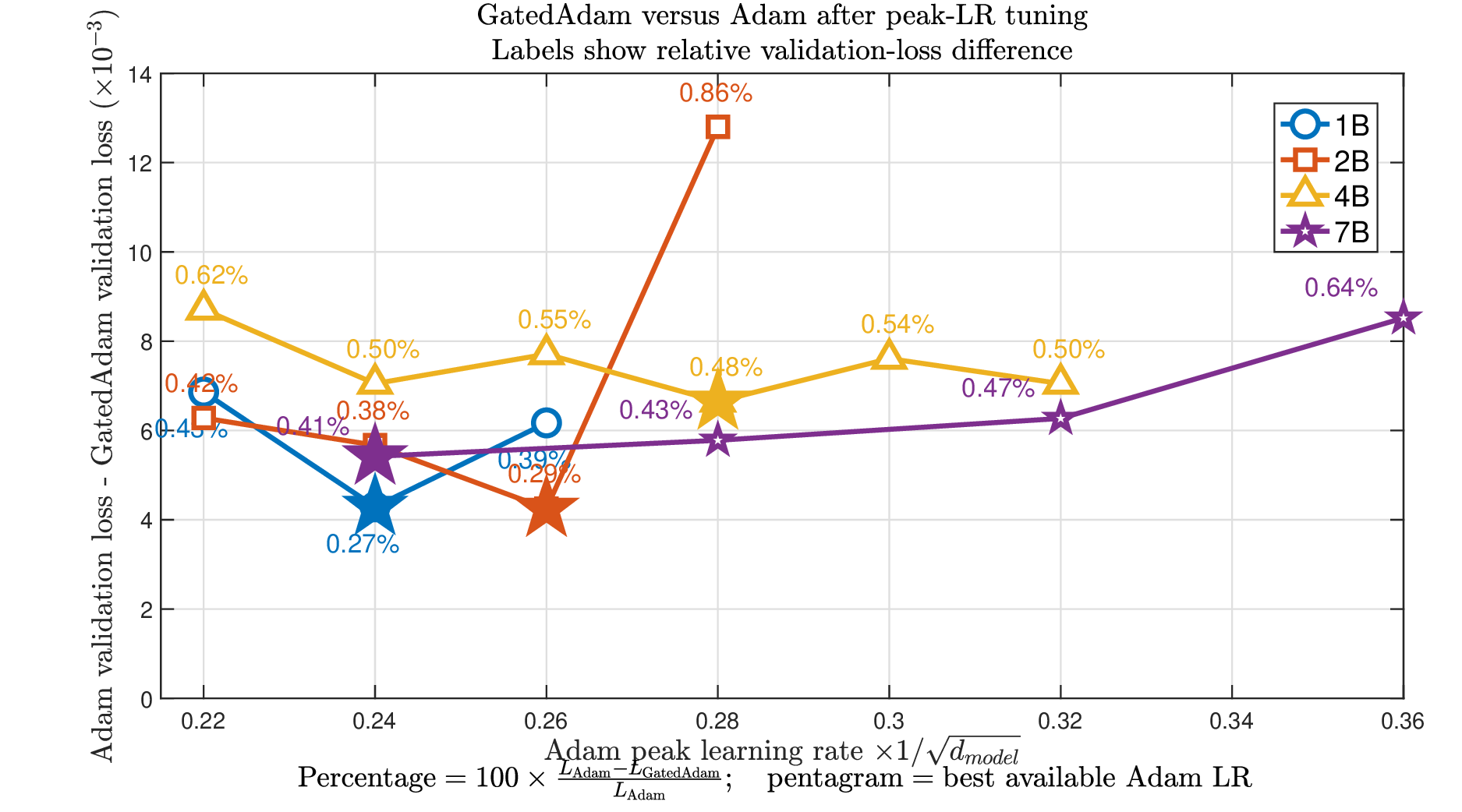}
    \caption{Validation loss comparison between nGPT trained with GatedAdam with $a=0.5$ and with
$a=1$ (Adam's gating) for models with 1B, 2B, 4B, and 7B total parameters. The $a=0.5$
setting uses a peak learning rate of $0.24/\sqrt{d_{\mathrm{model}}}$, whereas
the peak learning rate for $a=1$ is tuned separately for each model size. The
vertical axis reports $L_{a=1}-L_{a=0.5}$, so positive values indicate an
advantage for $a=0.5$. Percentage labels report
$100(L_{a=1}-L_{a=0.5})/L_{a=1}$, and filled pentagrams mark the best
available $a=1$ result for each model size.}
    \label{fig:gatedvsbase}
\end{figure}

\subsection{Experimental Results}

Figure~\ref{fig1:ladder} and
Table~\ref{tab:gpt_ngpt_gatedadam_losses} show the training and validation
losses obtained by GPT and nGPT for models ranging from 1B to 30B total
parameters. nGPT consistently achieves losses that are approximately 2.5--3.5\% 
lower. The results for the largest model demonstrate a larger performance gap between the two approaches. 

Figure~\ref{fig2:hist} shows the distributions of the bias-corrected Adam
second-moment scale $\sqrt{\hat v}$ used by GatedAdam at the end of training. Most parameter coordinates in these
distributions belong to MoE layers. The median $\sqrt{\hat v}$ decreases from $2.1\times10^{-9}$ for the
7B model ($d_{\mathrm{model}}=1280$) to $1.0\times10^{-9}$ for the
30B model ($d_{\mathrm{model}}=2048$). This is only a median-based estimate because 
the distribution is broad and the gate acts coordinate-wise. The distribution
of $\sqrt{\hat v}$ is asymmetric, with a relatively flat left tail that is
largely attributable to parameters in the input and output embeddings.

As shown in Figure~\ref{fig2:hist}, coordinates with small $\sqrt{\hat v}$ can receive gate values that are numerically close to zero. Setting $a$ to a smaller value, 
such as $0.5$, increases the gate values, and therefore the effective update
magnitudes, for coordinates with
$\sqrt{\hat v}<\epsilon_{\mathrm{gate}}$. We also investigated scheduling
$\epsilon_{\mathrm{gate}}$ as a function of model size or setting it based on
tensor statistics. The results of this investigation are outside the scope of
the current paper.

Figure~\ref{fig:lossplot} shows the convergence curves for the GPT and nGPT
versions of the MoE model with 30B total parameters and 3.23B active parameters
per token (the model corresponds to Nemotron-3 Nano). The shape of the GPT curve is strongly affected by the WSD
schedule. The nGPT curve is characterized by a slower initial decrease followed
by faster convergence later in training, consistent with the behavior observed
in \citet{loshchilov2024ngpt}. This behavior is consistent with parameter
normalization making the effective step size more directly controlled by the
learning rate schedule. Training and validation losses are evaluated on different data
distributions and therefore have different absolute scales; comparisons
should be made between GPT and nGPT within each split. The figure
shows that nGPT reaches the same training and validation losses using
approximately half as many training tokens as the GPT baseline trained with
AdamW.

\paragraph{Gating with $a=0.5$ versus $a=1$.}

The main GPT--nGPT comparison changes both the model parameterization and the
training recipe and therefore should not be interpreted as an isolated
comparison between Adam and GatedAdam. To estimate the contribution of gating within the normalized model,
Figure~\ref{fig:gatedvsbase} compares nGPT with $a=0.5$, using a peak learning
rate of $0.24/\sqrt{d_{\mathrm{model}}}$, against nGPT with $a=1$ (Adam's gating), for which
the peak learning rate is tuned separately at each model size. All other components of the nGPT training
recipe are applied in both cases, so the $a=1$ run is not a standalone Adam
baseline. The advantage of $a=0.5$ remains modest across model sizes and is
substantially smaller than the total gap between GPT and nGPT reported in
Table~\ref{tab:gpt_ngpt_gatedadam_losses}. This suggests that the overall
improvement arises primarily from the normalized parameterization and the
complete training recipe rather than from GatedAdam alone. At the matched validation-loss levels considered here, interpolation of the
loss--token curves indicates that the $a=0.5$ runs require approximately
15--20\% fewer training tokens than the best available $a=1$ runs. 

The ablation studies for Pre-Moment Tangent Projection and Post-Moment Exploration Noise are provided in Appendix \ref{appendix:projection} and Appendix \ref{appendix:noise}, respectively. Their effects on the validation loss are rather modest. 

\paragraph{SwiGLU vs Squared ReLU.}

\begin{figure}[!t]
    \centering
    \includegraphics[width=0.98\textwidth]{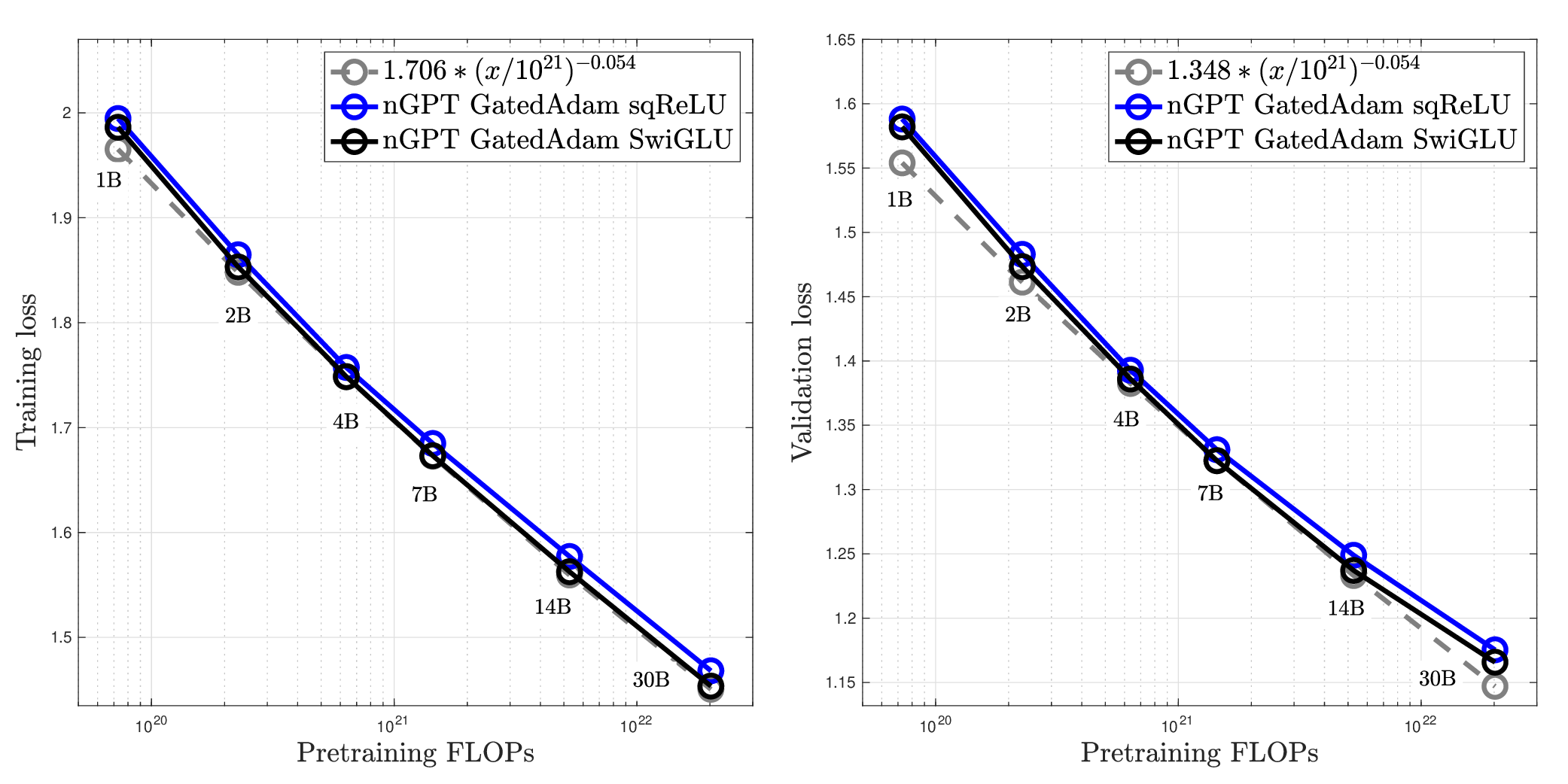}
    \caption{Training (left) and validation (right) loss versus pretraining FLOPs for normalized hybrid Mamba-2–Transformer MoE models trained with GatedAdam, using squared ReLU (blue) or SwiGLU (black). Markers indicate total model sizes from 1B to 30B parameters. Gray dashed curves show reference power laws \(L=a(C/10^{21})^{-0.054}\), where \(C\) denotes pretraining FLOPs and \(a=1.706\) for training and \(a=1.348\) for validation. SwiGLU yields lower training and validation losses at every model size.}
    \label{fig:ladder_swiglu}
\end{figure}

The original experiments with nGPT \citep{loshchilov2024ngpt} used
SwiGLU \citep{shazeer2020glu} as the activation function in MLP blocks.
However, in this paper, for the sake of comparability, we adopted the
Nemotron-3 architecture  which uses Squared ReLU. In this section, we
evaluate this choice and compare SwiGLU (implemented as in
\cite{loshchilov2024ngpt}) to sqReLU. Figure~\ref{fig:ladder_swiglu}
shows the results of this comparison together with the scaling curves
for both training and validation losses. The latter curves are not
obtained by fitting all existing data points but represent our estimates
of the actual scaling. For instance, smaller models approach these
envelope curves from above, with agreement starting from the 4B models.
Validation loss curves start to depart from the trend for
14B models trained for 560B tokens. We found the reason for this: our
dataset contained about 500B unique tokens, with some data undergoing
more passes than average. Thus, for the 30B model trained on 1000B tokens,
most tokens have been seen at least twice, which leads to worse
generalization than using fresh tokens (i.e., one repeated token is worth
less than one fresh token). Training curves for SwiGLU show almost
perfect scaling from 4B to 30B models (see Figure~\ref{fig:ladder_swiglu_normalized} in the Appendix).
Table~\ref{tab:ngpt_sqrelu_swiglu_losses} shows that the gap between
SwiGLU and sqReLU grows and reaches a 1\% difference in training loss.
Although SwiGLU has the same parameter count and FLOPs cost as sqReLU,
in practice it may be more expensive to run. In our experiments, SwiGLU
was up to 5\% slower than sqReLU. However, the 1\% loss difference could
translate into about 20--25\% acceleration of pretraining. Thus, its net
advantage of 15--20\% and the observed increase in the gap with the
compute budget suggest its adoption as the baseline activation function
in nGPT.

\begin{table}
\centering
\setlength{\tabcolsep}{3pt}
\caption{nGPT GatedAdam losses with sqReLU and SwiGLU.
Differences are sqReLU minus SwiGLU; percentages are relative
to sqReLU. Differences are computed before rounding.}
\label{tab:ngpt_sqrelu_swiglu_losses}
\begin{tabular}{@{}rc*{8}{r}@{}}
\toprule
\shortstack{Model\\Size} & Tokens
& \multicolumn{4}{c}{Training loss}
& \multicolumn{4}{c}{Validation loss} \\
\cmidrule(lr){3-6}
\cmidrule(lr){7-10}
& & sqReLU & SwiGLU & $\Delta$ & $\Delta$ (\%)
  & sqReLU & SwiGLU & $\Delta$ & $\Delta$ (\%) \\
\midrule
1B & 88B
& 1.9946 & 1.9863 & 0.0083 & 0.42\%
& 1.5881 & 1.5820 & 0.0061 & 0.38\% \\
2B & 140B
& 1.8648 & 1.8531 & 0.0117 & 0.63\%
& 1.4829 & 1.4734 & 0.0095 & 0.64\% \\
4B & 215B
& 1.7573 & 1.7486 & 0.0087 & 0.50\%
& 1.3928 & 1.3859 & 0.0069 & 0.49\% \\
7B & 310B
& 1.6849 & 1.6732 & 0.0117 & 0.69\%
& 1.3310 & 1.3224 & 0.0086 & 0.65\% \\
14B & 560B
& 1.5772 & 1.5626 & 0.0146 & 0.92\%
& 1.2489 & 1.2371 & 0.0118 & 0.94\% \\
30B & 1000B
& 1.4682 & 1.4537 & 0.0146 & 0.99\%
& 1.1754 & 1.1658 & 0.0097 & 0.82\% \\
\bottomrule
\end{tabular}
\end{table}

\section{Conclusion}
\label{section_discussionconclusion}

This paper presents a practical recipe for training nGPT with modern hybrid
Mixture-of-Experts models and demonstrates a substantial improvement in data
efficiency. Further work may extend the evaluation to larger models and
investigate hyperparameter scaling rules.

\section*{Acknowledgments}

We thank Roger Waleffe for providing the source code and logs for the baseline
GPT experiments with AdamW. We thank Mikail Khona, Kwangjun Ahn, and Roger
Waleffe for providing the scaling ladder used in our experiments. Finally, we
thank Mostofa Patwary, Mohammad Shoeybi, and the entire Nemotron Team
\citep{blakeman2025nemotron} for their contributions to the development of the
Nemotron open models.

\bibliography{iclr2025_conference}
\bibliographystyle{iclr2025_conference}

\clearpage
\appendix

\section{GatedAdamW's gate without an explicit sigmoid}
\label{appendix:equivalentform}

The sigmoid-of-log gate has an equivalent power-law form. For each coordinate
with $d_{t,i}>0$,
\begin{align}
    \gamma_{t,i}
    &=
    \frac{1}{
        1+\exp\!\left(
            -a\log\frac{d_{t,i}}{\epsilon_{\mathrm{gate}}}
        \right)
    } \\
    &=
    \frac{1}{
        1+\left(
            \epsilon_{\mathrm{gate}}/d_{t,i}
        \right)^a
    } \\
    &=
    \frac{d_{t,i}^{\,a}}
         {d_{t,i}^{\,a}+\epsilon_{\mathrm{gate}}^{\,a}} .
\end{align}
Thus, in vector notation,
\begin{align}
    \bm{\gamma}_t
    =
    \frac{\vd_t^{\,a}}
         {\vd_t^{\,a}+\epsilon_{\mathrm{gate}}^{\,a}},
\end{align}
where powers, divisions, and additions are applied coordinate-wise. The
adaptive part of the update can therefore be written without an explicit
sigmoid as
\begin{align}
    \bm{\gamma}_t
    \odot
    \frac{\hat{\vm}_t}{\vd_t}
    =
    \frac{\vd_t^{\,a-1}}
         {\vd_t^{\,a}+\epsilon_{\mathrm{gate}}^{\,a}}
    \odot
    \hat{\vm}_t .
\end{align}
This form makes the power-law dependence on $\vd_t$ explicit. In particular,
when $a=1$,
\begin{align}
    \bm{\gamma}_t
    \odot
    \frac{\hat{\vm}_t}{\vd_t}
    =
    \frac{\hat{\vm}_t}
         {\vd_t+\epsilon_{\mathrm{gate}}}.
\end{align}
Hence, if $\epsilon_{\mathrm{num}}=0$ and
$\epsilon_{\mathrm{gate}}=\epsilon$, the adaptive update reduces to the
standard AdamW update.

\section{Pre-Moment Tangent Projection}
\label{appendix:projection}

For parameter vectors constrained to the unit hypersphere, only the tangent
component of the gradient contributes to the first-order change after
normalization. We therefore optionally project the gradient onto the tangent
space before the first- and second-moment updates of GatedAdamW. Tangent-space
gradient projection is an established operation in Riemannian optimization
methods for scale-invariant parameters
\citep{cho2017riemannian}. We do not claim the projection itself as a novel
contribution; we use the descriptive term \emph{Pre-Moment Tangent Projection}
to emphasize that it is applied before both optimizer moments are updated and
to provide a concise reference within this paper. For each normalized vector
$\vw$ and its gradient $\vg$, we replace
\begin{align}
    \vg
    \leftarrow
    \vg
    -
    \vw
    \frac{\langle \vw,\vg\rangle}
         {\max(\langle \vw,\vw\rangle,\epsilon_{\mathrm{proj}})}.
\end{align}
When $\|\vw\|_2=1$, this projection simply removes the radial component
$\langle \vw,\vg\rangle\vw$, and the safeguard
$\epsilon_{\mathrm{proj}}=10^{-12}$ is inactive. GatedAdamW's moment estimates
are then updated using the projected gradient. Thus, the moments do not
accumulate radial gradient components that would subsequently be discarded by
normalization. Unlike a fully Riemannian Adam update, this procedure does not
parallel-transport the first-moment estimate after the parameter update.

In the distributed implementation, the dot products and squared norms are
computed over each complete normalized row or column vector, even when the
vector is split across distributed optimizer shards. The required statistics
are all-reduced before the local gradient shards are modified.

\begin{figure}
    \centering
    \includegraphics[width=0.8\textwidth]{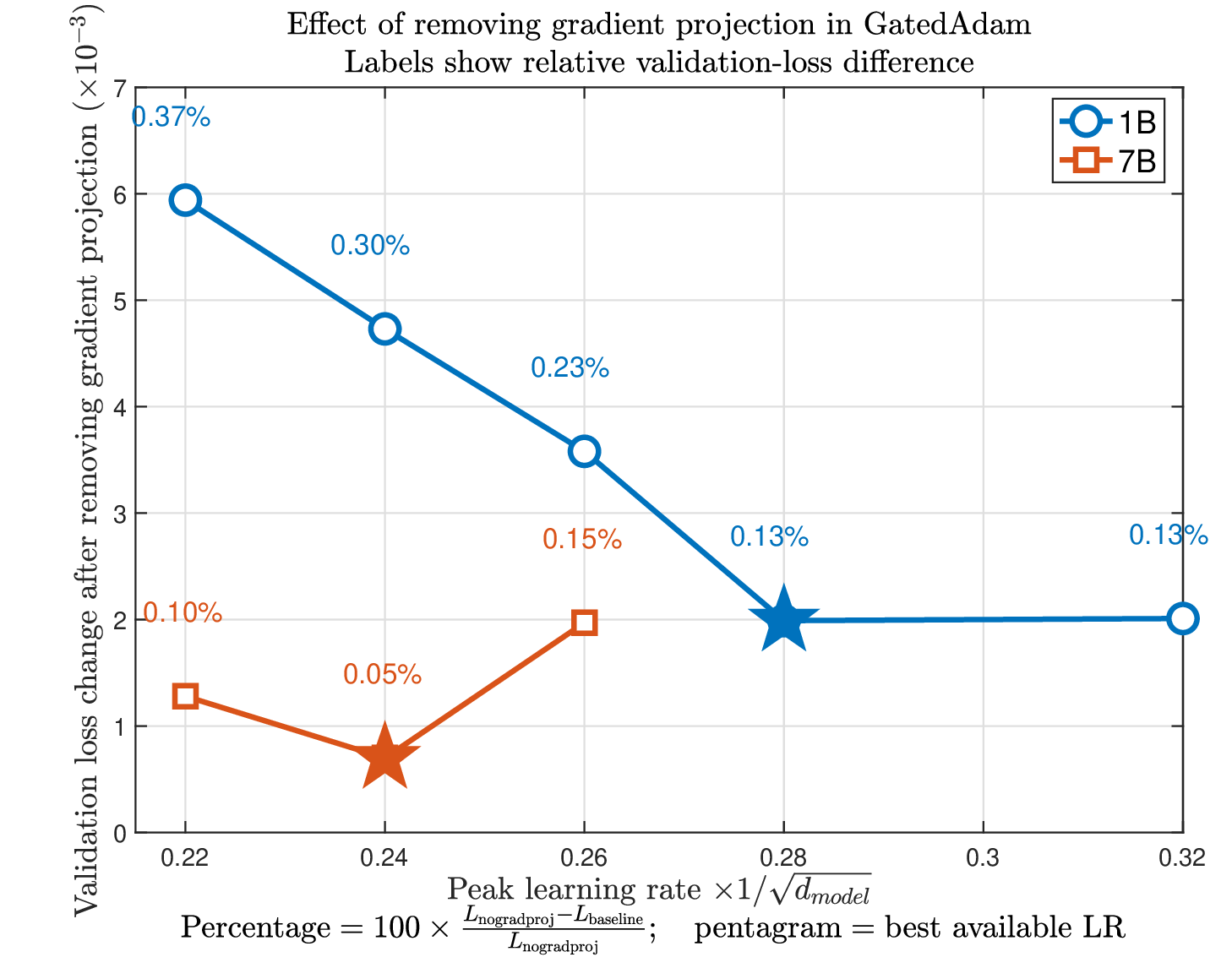}
    \caption{Validation loss change after removing Pre-Moment Tangent Projection from nGPT for models with 1B and 7B total parameters. The default nGPT uses a peak learning rate of $0.24/\sqrt{d_{\mathrm{model}}}$, whereas for the case without projections the peak learning rate is tuned separately for each model size. }
    \label{fig:gproj}
\end{figure}

Figure~\ref{fig:gproj} illustrates the effect of Pre-Moment
Tangent Projection. It slightly improves the final validation loss for the 1B model, while its effect is smaller for the 7B model.

\section{Second-Moment Growth Clipping (SMGC)}
\label{appendix:clipping}

To limit the effect of isolated gradient spikes on the moment estimates, we
optionally clip each scalar gradient coordinate using only its own
second-moment history. Let $v_i$ denote the existing bias-uncorrected
second-moment state for coordinate $i$, and let $k$ be the number of completed
optimizer steps. We define
\begin{align}
    \bar v_i
    &=
    \max\left(
        v_i,\,
        \left(1-\beta_2^k\right)s_{\min}^2
    \right), \\
    C_R
    &=
    \sqrt{
        \frac{R-\beta_2}
             {1-\beta_2}
    },
\end{align}
where $s_{\min}$ is a floor expressed in bias-corrected
$\sqrt{\hat v}$ units and $R\geq 1$ controls the maximum permitted one-step
growth relative to $\bar v_i$. The gradient is then replaced coordinate-wise by
\begin{align}
    g_i
    \leftarrow
    \operatorname{sign}(g_i)
    \min\left(
        |g_i|,
        C_R\sqrt{\bar v_i}
    \right).
\end{align}
Because $v_i\leq\bar v_i$, the subsequent second-moment update satisfies
\begin{align}
    v_i^{+}
    =
    \beta_2 v_i
    +
    (1-\beta_2)g_i^2
    \leq
    R\bar v_i.
\end{align}
For coordinates above the floor, $\bar v_i=v_i$, and the rule directly bounds
the growth of the second moment by $v_i^{+}\leq Rv_i$. For coordinates below
the floor, the bound is instead defined relative to the floor. The clipped
gradient is used to update both the first- and second-moment estimates.

Clipping is applied after gradient unscaling and, when enabled, Pre-Moment
Tangent Projection, but before the first- and second-moment updates of
GatedAdamW. It may be activated only after a prescribed number of optimizer
steps to allow the second-moment statistics to initialize. Unlike global norm
clipping, SMGC uses no statistics from other coordinates or parameters.
Because it operates coordinate-wise, however, it may change the direction of a
multidimensional gradient.

SMGC is closely related to the spike-aware gradient clipping used in SPAM
\citep{huang2025spam}, which identifies unusually large gradients relative to
AdamW's running second-moment estimate and rescales them before they enter the
moment updates. Unlike SPAM, we do not periodically reset the first- and
second-moment states. Instead, the clipping threshold is parameterized
directly through the maximum permitted one-step second-moment growth factor
$R$, together with a floor for coordinates with little second-moment history
and an optional delayed activation period.

\begin{figure}
    \centering
    \includegraphics[width=0.8\textwidth]{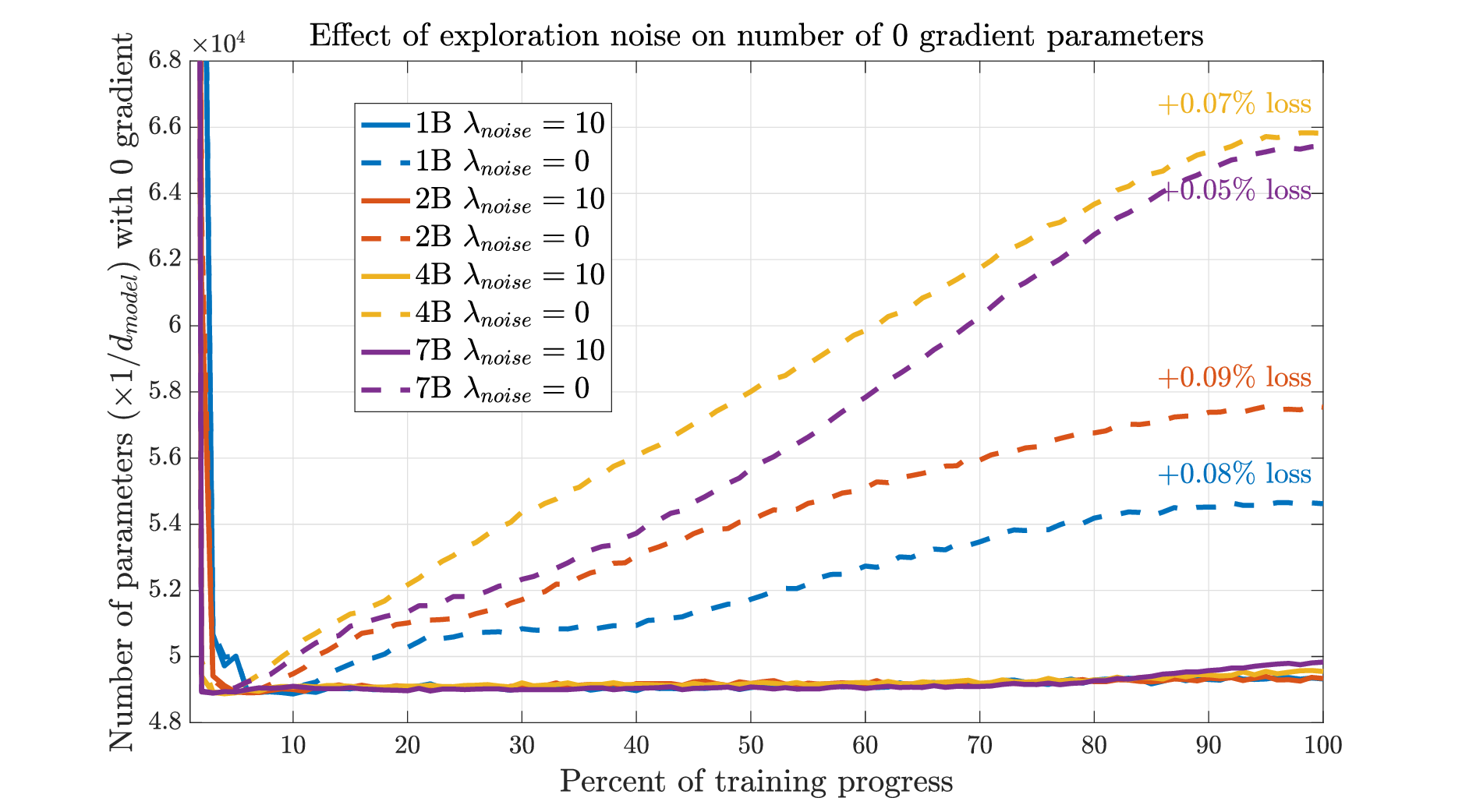}
    \caption{Evolution of the number of parameter coordinates with exactly zero gradient for models with 1B, 2B, 4B, and 7B total parameters. Solid curves show the default nGPT runs with Post-Moment Exploration Noise (PMEN, $\lambda_{\mathrm{noise}}=10$), while dashed curves show otherwise matched runs without exploration noise ($\lambda_{\mathrm{noise}}=0$). The zero-gradient count is divided by $d_{\mathrm{model}}$ to facilitate comparison across model widths. The annotations show the validation loss change. }
    \label{fig:numzeros}
\end{figure}

\section{Post-Moment Exploration Noise (PMEN)}
\label{appendix:noise}

We optionally inject a parameter-space perturbation after the optimizer moment
update. Let $\widetilde{\vw}_{j,t}$ denote the post-GatedAdamW candidate for a
parameter vector $\vw_j\in\mathbb{R}^{d_j}$, and let $\alpha_{j,t}$ denote the current scheduled learning rate of parameter group $j$, including any parameter-specific multiplier and warmup. We associate each coordinate with an
activity scale
\begin{align}
    s_{j,i,t}
    =
    \sqrt{v_{j,i,t}},
\end{align}
where $v_{j,i,t}$ is Adam's raw, bias-uncorrected second-moment state after the
current moment update.

Let $\psi(s;\tau_v)\geq0$ be a non-increasing noise profile and let $c_{j,t}$
be a reference coordinate scale. At selected optimizer steps, we draw
independent Rademacher signs $\xi_{j,i,t}\in\{-1,+1\}$ and apply
\begin{align}
    \widetilde{\vw}_{j,t}
    \leftarrow
    \widetilde{\vw}_{j,t}
    +
    \alpha_{j,t}
    \lambda_{\mathrm{noise}}
    c_{j,t}
    \left(
        \boldsymbol{\psi}_{j,t}
        \odot
        \vxi_{j,t}
    \right),
\end{align} 
where
\begin{align}
    [\boldsymbol{\psi}_{j,t}]_i
    =
    \psi(s_{j,i,t};\tau_v).
\end{align}
Here, $\lambda_{\mathrm{noise}}$ controls the overall perturbation strength,
$\psi$ distributes it across second-moment scales, and $c_{j,t}$ specifies its
reference scale.

A sparse hard-threshold variant is obtained with
\begin{align}
    \psi_{\mathrm{step}}(s;\tau_v)
    =
    \mathbb{I}[s\leq\tau_v].
\end{align}
Setting $\tau_v=0$ restricts the perturbation to coordinates whose
second-moment state is exactly zero. A smooth alternative is
\begin{align}
    \psi_b(s;\tau_v)
    =
    \frac{1}
         {1+\left(s/\tau_v\right)^b},
    \qquad
    \tau_v>0,
\end{align}
where $b>0$ controls the sharpness of the transition. As
$b\rightarrow\infty$, this profile approaches the hard-threshold rule, apart
from its value exactly at $s=\tau_v$.

One possible reference scale is
\begin{align}
    c_{j,t}
    =
    \frac{1}{\sqrt{d_j}},
\end{align}
which normalizes the perturbation at the vector level. An Adam-relative
alternative is
\begin{align}
    c_{j,t}
    =
    c_{\mathrm{Adam}}
    =
    \sqrt{
        \frac{1-\beta_1}
             {1+\beta_1}
    },
\end{align}
which approximates the coordinate-wise root-mean-square normalized Adam update
under stationary random gradients. In this parameterization,
$\lambda_{\mathrm{noise}}$ measures the perturbation relative to the
coordinate-wise stochastic update scale of Adam.

For the hard-threshold profile, if $k_j$ coordinates are selected, then
\begin{align}
    \left\|
    \Delta\vw_{j,t}^{\mathrm{noise}}
\right\|_2^2
=
\alpha_{j,t}^2
\lambda_{\mathrm{noise}}^2
c_{j,t}^2
k_j.
\end{align}
For a unit-normalized vector and a sufficiently small perturbation, the
first-order angular displacement is bounded by this Euclidean perturbation
norm.

The perturbation is applied after the Adam moment update and the GatedAdamW
parameter correction, and therefore does not enter either moment estimate. 
For normalized non-embedding parameters covered by ASC, the Angular Step Cap is subsequently applied to the combined optimizer and noise displacement, followed by the usual nGPT normalization. PMEN may be delayed, applied periodically, and enabled or
disabled for embedding and output matrices. 

Figure~\ref{fig:numzeros} illustrates the effect of PMEN. Solid curves denote
GatedAdam runs with $\lambda_{\mathrm{noise}}=10$, the value used in our
baseline experiments, whereas dashed curves denote otherwise matched
noise-free runs with $\lambda_{\mathrm{noise}}=0$. The runs with PMEN achieve
slightly lower validation loss, although the improvement is small and does not
exceed $0.1\%$. Without PMEN, the number of parameter coordinates with exactly
zero gradients increases with model size even after normalization by
$d_{\mathrm{model}}$. With PMEN, the normalized count remains approximately
stable, with only a small increase for the largest model.

At the end of training, approximately $1.4\%$ of the input-embedding parameter
coordinates in the 4B model have $\sqrt{\hat v}=0$, indicating that their Adam
second-moment states have never received a nonzero gradient contribution. For
the input embeddings, this is consistent with some token IDs never appearing
in the training data. Approximately $0.7\%$ of the routed-expert parameter
coordinates also have $\sqrt{\hat v}=0$. PMEN is intended to reduce the
likelihood that weakly updated parameter coordinates become permanently
inactive and to preserve their opportunity to receive useful gradients later
in training. Although PMEN reduces both the number of coordinates with zero
gradients and the number with $\sqrt{\hat v}=0$, we do not yet observe a
substantial improvement in final loss.

\section{Additional SwiGLU Scaling Results}

\begin{figure}[!t]
    \centering
    \includegraphics[width=0.98\textwidth]{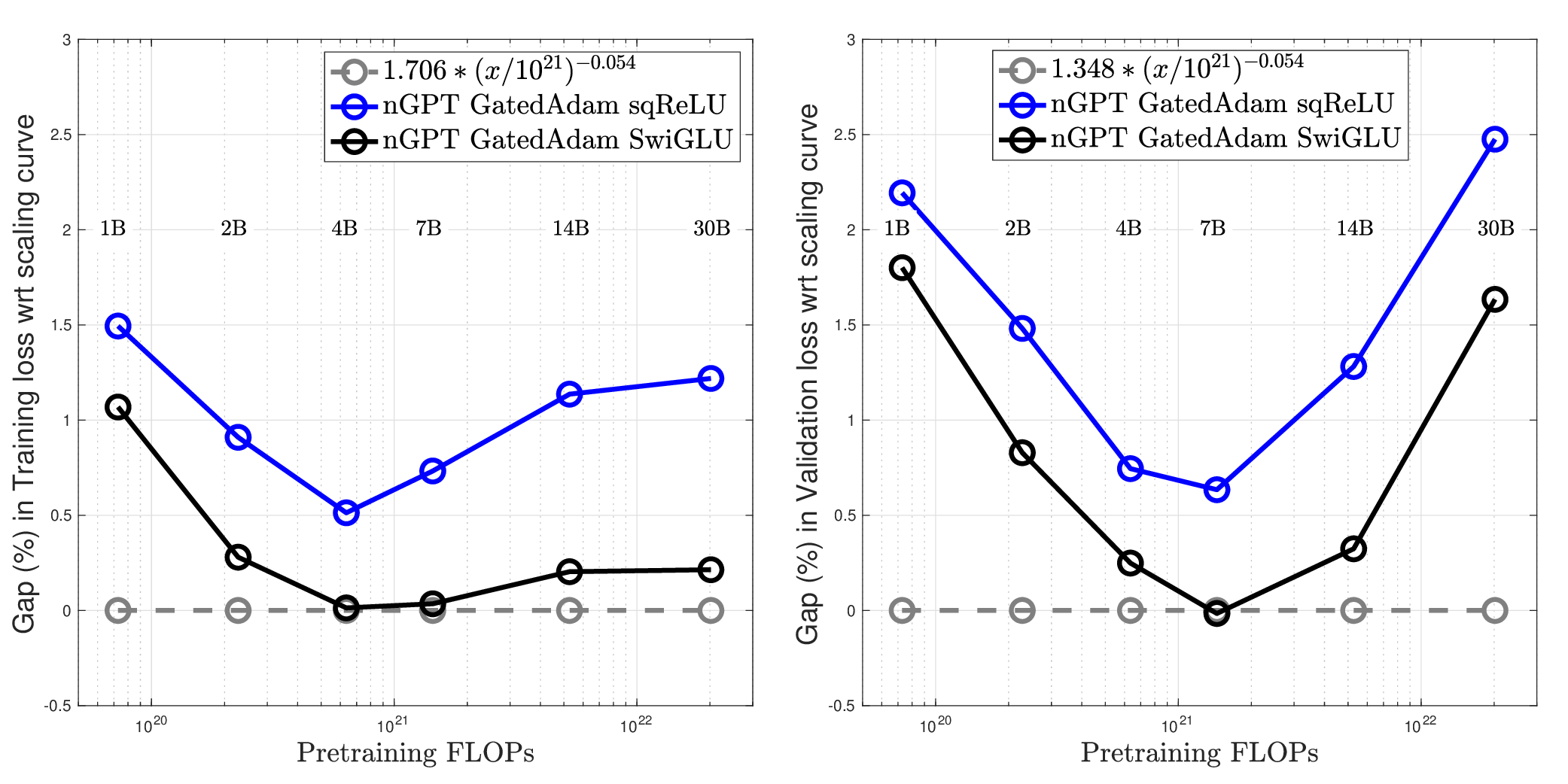}
    \caption{Percentage deviations of training (left) and validation (right) losses from reference scaling curves for normalized hybrid Mamba-2–Transformer MoE models trained with GatedAdam, using squared ReLU (blue) or SwiGLU (black). The gap is \(100(L/L_{\mathrm{ref}}-1)\), where \(L_{\mathrm{ref}}=a(C/10^{21})^{-0.054}\), \(C\) denotes pretraining FLOPs, and \(a=1.706\) for training and \(a=1.348\) for validation. The gray dashed line marks zero deviation. Markers correspond to total model sizes from 1B to 30B parameters. SwiGLU tracks the reference curves more closely at every scale, while validation deviations increase at the 14B and 30B scales. The deviation of the validation loss curve at 30B is primarily due the use of repeated tokens (the total training budget was 1000B tokens but the dataset has only 500B unique tokens).}
    \label{fig:ladder_swiglu_normalized}
\end{figure}
\vspace{0.25cm}
\end{document}

%% file: math_commands.tex
\usepackage{amsmath,amsfonts,bm}

\def\eqref#1{equation~\ref{#1}}

\def\1{\bm{1}}

\def\vd{{\bm{d}}}

\def\vg{{\bm{g}}}
\def\vh{{\bm{h}}}

\def\vk{{\bm{k}}}

\def\vm{{\bm{m}}}

\def\vq{{\bm{q}}}

\def\vs{{\bm{s}}}

\def\vu{{\bm{u}}}
\def\vv{{\bm{v}}}
\def\vw{{\bm{w}}}

\def\vz{{\bm{z}}}

\def\mW{{\bm{W}}}

\DeclareMathAlphabet{\mathsfit}{\encodingdefault}{\sfdefault}{m}{sl}
\SetMathAlphabet{\mathsfit}{bold}{\encodingdefault}{\sfdefault}{bx}{n}



%% file: iclr2025_conference.bib
@inproceedings{wang2020understanding,
  title={Understanding contrastive representation learning through alignment and uniformity on the hypersphere},
  author={Wang, Tongzhou and Isola, Phillip},
  booktitle={ ICML},
  year={2020},
}

@article{mettes2019hyperspherical,
  title={Hyperspherical prototype networks},
  author={Mettes, Pascal and Van der Pol, Elise and Snoek, Cees},
  journal={NeurIPS},
  year={2019}
}

@article{xu2018spherical,
  title={Spherical latent spaces for stable variational autoencoders},
  author={Xu, Jiacheng and Durrett, Greg},
  journal={ arXiv:1808.10805},
  year={2018}
}

@inproceedings{wang2017normface,
  title={Normface: L2 hypersphere embedding for face verification},
  author={Wang, Feng and Xiang, Xiang and Cheng, Jian and Yuille, Alan Loddon},
  booktitle={Proc. of the 25th ACM nternational conference on Multimedia},
  year={2017}
}

@inproceedings{loshchilov2017decoupled,
  title={Decoupled Weight Decay Regularization},
  author={Loshchilov, Ilya and Hutter, Frank},
  booktitle={ICLR},
  year={2019}
}

@article{kodryan2022training,
  title={Training scale-invariant neural networks on the sphere can happen in three regimes},
  author={Kodryan, Maxim and Lobacheva, Ekaterina and Nakhodnov, Maksim and Vetrov, Dmitry P},
  journal={NeurIPS},
  year={2022}
}

@article{kosson2023rotational,
  title={Rotational equilibrium: How weight decay balances learning across neural networks},
  author={Kosson, Atli and Messmer, Bettina and Jaggi, Martin},
  journal={ arXiv:2305.17212},
  year={2023}
}

@article{salimans2016weight,
  title={Weight normalization: A simple reparameterization to accelerate training of deep neural networks},
  author={Salimans, Tim and Kingma, Durk P},
  journal={NeurIPS},
  year={2016}
}

@article{franke2023cpr,
      title={Constrained Parameter Regularization}, 
      author={Jörg K. H. Franke and Michael Hefenbrock and Gregor Koehler and Frank Hutter},
      year={2023},
      journal={ arXiv:2311.09058},
}

@article{shazeer2020glu,
  title={Gated Linear Units (GLU)},
  author={Noam Shazeer},
  year={2020},
  journal={ arXiv:2002.05202}
}

@inproceedings{liu2018decoupled,
  title={Decoupled networks},
  author={Liu, Weiyang and Liu, Zhen and Yu, Zhiding and Dai, Bo and Lin, Rongmei and Wang, Yisen and Rehg, James M and Song, Le},
  booktitle={Proceedings of the IEEE Conference on Computer Vision and Pattern Recognition},
  pages={2771--2779},
  year={2018}
}

@article{liu2017deep,
  title={Deep hyperspherical learning},
  author={Liu, Weiyang and Zhang, Yan-Ming and Li, Xingguo and Yu, Zhiding and Dai, Bo and Zhao, Tuo and Song, Le},
  journal={Advances in neural information processing systems},
  volume={30},
  year={2017}
}

@article{loshchilov2023weight,
  title={Weight norm control},
  author={Loshchilov, Ilya},
  journal={arXiv preprint arXiv:2311.11446},
  year={2023}
}

@inproceedings{karras2024analyzing,
  title={Analyzing and improving the training dynamics of diffusion models},
  author={Karras, Tero and Aittala, Miika and Lehtinen, Jaakko and Hellsten, Janne and Aila, Timo and Laine, Samuli},
  booktitle={Proceedings of the IEEE/CVF Conference on Computer Vision and Pattern Recognition},
  pages={24174--24184},
  year={2024}
}

@inproceedings{liu2021learning,
  title={Learning by turning: Neural architecture aware optimisation},
  author={Liu, Yang and Bernstein, Jeremy and Meister, Markus and Yue, Yisong},
  booktitle={International Conference on Machine Learning},
  pages={6748--6758},
  year={2021},
  organization={PMLR}
}

@article{loshchilov2024ngpt,
  title={n{GPT}: Normalized Transformer with Representation Learning on the Hypersphere},
  author={Loshchilov, Ilya and Hsieh, Cheng-Ping and Sun, Simeng and Ginsburg, Boris},
  journal={arXiv e-prints},
  pages={arXiv--2410},
  year={2024}
}

@article{blakeman2025nemotron,
  title={Nemotron 3 Nano: Open, Efficient Mixture-of-Experts Hybrid Mamba-Transformer Model for Agentic Reasoning},
  author={Blakeman, Aaron and Grattafiori, Aaron and Basant, Aarti and Gupta, Abhibha and Khattar, Abhinav and Renduchintala, Adi and Vavre, Aditya and Shukla, Akanksha and Bercovich, Akhiad and Ficek, Aleksander and others},
  journal={arXiv preprint arXiv:2512.20848},
  year={2025}
  }

@article{loshchilov2016sgdr,
  title={{SGDR}: Stochastic gradient descent with warm restarts},
  author={Loshchilov, Ilya and Hutter, Frank},
  journal={arXiv preprint arXiv:1608.03983},
  year={2016}
}

@article{waleffe2024empirical,
  title={An empirical study of mamba-based language models},
  author={Waleffe, Roger and Byeon, Wonmin and Riach, Duncan and Norick, Brandon and Korthikanti, Vijay and Dao, Tri and Gu, Albert and Hatamizadeh, Ali and Singh, Sudhakar and Narayanan, Deepak and others},
  journal={arXiv preprint arXiv:2406.07887},
  year={2024}
}

@inproceedings{henry2020querykey,
  title     = {Query-Key Normalization for Transformers},
  author    = {Henry, Alex and Dachapally, Prudhvi Raj and Pawar, Shubham Shantaram and Chen, Yuxuan},
  booktitle = {Findings of the Association for Computational Linguistics: EMNLP 2020},
  pages     = {4246--4253},
  year      = {2020},
  publisher = {Association for Computational Linguistics},
  doi       = {10.18653/v1/2020.findings-emnlp.379}
}

@inproceedings{nguyen2019transformers,
  title     = {Transformers without Tears: Improving the Normalization of Self-Attention},
  author    = {Nguyen, Toan Q. and Salazar, Julian},
  booktitle = {Proceedings of the 16th International Conference on Spoken Language Translation},
  year      = {2019},
  publisher = {Association for Computational Linguistics}
}

@inproceedings{dao2024mamba2,
  title     = {Transformers are {SSM}s: Generalized Models and Efficient Algorithms Through Structured State Space Duality},
  author    = {Dao, Tri and Gu, Albert},
  booktitle = {Proceedings of the 41st International Conference on Machine Learning},
  series    = {Proceedings of Machine Learning Research},
  volume    = {235},
  pages     = {10041--10071},
  year      = {2024},
  publisher = {PMLR}
}

@misc{name2026personal,
  author = {Mikail Khona and Kwangjun Ahn and Roger Waleffe and Mostofa Patwary and Mohammad
Shoeybi},
  title        = {Personal communication},
  year         = {2026},
  note         = {Personal communication}
}

@article{hu2024minicpm,
  title={Minicpm: Unveiling the potential of small language models with scalable training strategies},
  author={Hu, Shengding and Tu, Yuge and Han, Xu and He, Chaoqun and Cui, Ganqu and Long, Xiang and Zheng, Zhi and Fang, Yewei and Huang, Yuxiang and Zhao, Weilin and others},
  journal={arXiv preprint arXiv:2404.06395},
  year={2024}
}

@article{xing2018walk,
  title={A walk with sgd},
  author={Xing, Chen and Arpit, Devansh and Tsirigotis, Christos and Bengio, Yoshua},
  journal={arXiv preprint arXiv:1802.08770},
  year={2018}
}

@article{huang2025spam,
  title   = {{SPAM}: Spike-Aware Adam with Momentum Reset for Stable
             {LLM} Training},
  author  = {Huang, Tianjin and Zhu, Ziquan and Jin, Gaojie and Liu, Lu
             and Wang, Zhangyang and Liu, Shiwei},
  journal = {arXiv preprint arXiv:2501.06842},
  year    = {2025}
}

@article{cho2017riemannian,
  title={Riemannian approach to batch normalization},
  author={Cho, Minhyung and Lee, Jaehyung},
  journal={Advances in Neural Information Processing Systems},
  volume={30},
  year={2017}
}
